\documentclass[twocolumn,10pt,aps,superscriptaddress,showkeys,noeprint]{revtex4-2}
\pdfoutput=1
\usepackage{style}

\usepackage[capitalise]{cleveref}

\begin{document}

\title{American cultural regions mapped through the lexical analysis of social media}

\author{Thomas Louf\footnote{thomaslouf@ifisc.uib-csic.es}}
\email{thomaslouf@ifisc.uib-csic.es}
\affiliation{Instituto de F\'{\i}sica Interdisciplinar y Sistemas Complejos IFISC (UIB-CSIC), Palma de Mallorca, Spain}

\author{Bruno Gon\c calves}
\affiliation{ISI Foundation, Turin, Italy}

\author{Jos\'e J. Ramasco}
\affiliation{Instituto de F\'{\i}sica Interdisciplinar y Sistemas Complejos IFISC (UIB-CSIC), Palma de Mallorca, Spain}

\author{David S\'{a}nchez\footnote{david.sanchez@uib.es}}
\email{david.sanchez@uib.es}
\affiliation{Instituto de F\'{\i}sica Interdisciplinar y Sistemas Complejos IFISC (UIB-CSIC), Palma de Mallorca, Spain}

\author{Jack Grieve}
\affiliation{Department of English Language and Linguistics, University of Birmingham, Birmingham, UK}
\affiliation{The Alan Turing Institute, London, UK}


\date{\today}


  \begin{abstract}
    Cultural areas represent a useful concept that cross-fertilizes diverse fields
    in social sciences. Knowledge of how humans organize and relate their ideas and behavior within a society helps to understand their actions and attitudes towards different issues. However, the selection of common traits that shape a cultural area is somewhat arbitrary. What is needed is a method that can leverage the massive amounts of data coming online, especially through social media, to identify cultural regions without ad-hoc assumptions, biases or prejudices. This work takes a crucial step in this direction by introducing a method to infer cultural regions based on the automatic analysis of large datasets from microblogging posts. The approach presented here is based on the principle that cultural affiliation can be inferred from the topics that people discuss among themselves. Specifically, regional variations in written discourse are measured in American social media. From the frequency distributions of content words in geotagged Tweets, the regional hotspots of words' usage are found, and from there, principal components of regional variation are derived. Through a hierarchical clustering of the data in this lower-dimensional space, this method yields clear cultural areas and the topics of discussion that define them. It uncovers a manifest North-South separation, which is primarily influenced by the African American culture, and further contiguous (East-West) and non-contiguous divisions
    that provide a comprehensive picture of today's cultural areas in the US.
  \end{abstract}

\clearpage

\maketitle


\section*{Introduction}

Cultural identity is an elusive notion because it depends on a wide range of different
cultural factors---including politics, religion, ethnicity, economics, and art, among
countless other examples---which will generally differ across individuals, with the
cultural background of every individual ultimately being unique. Nevertheless, individuals from the same region can generally be expected to share some
cultural traits, reflecting the shared cultural values and practices associated with the
region~\citep{broek1973geography}. Identifying the cultural regions of
a nation---regions whose populations are characterized by relative cultural
homogeneity compared to the populations of other regions within the nation---is very valuable information across a wide range of domains. For example, it is
important for governments to understand geographical variation in the values of their
population so as to better meet their educational, social, and welfare needs. Similarly,
from an economic standpoint, it is important to identify where certain services and
products are most required and how best to engage with populations in different regions
of the country. In general, defining the cultural regions of a nation is therefore a
crucial part of understanding the complex landscape of human behavior that nation
encompasses, providing an accessible and broad classification of the populations of a
country~\citep{lane2016culture}.

\begin{figure*}
  \centering
  \includegraphics[width=0.8\textwidth]{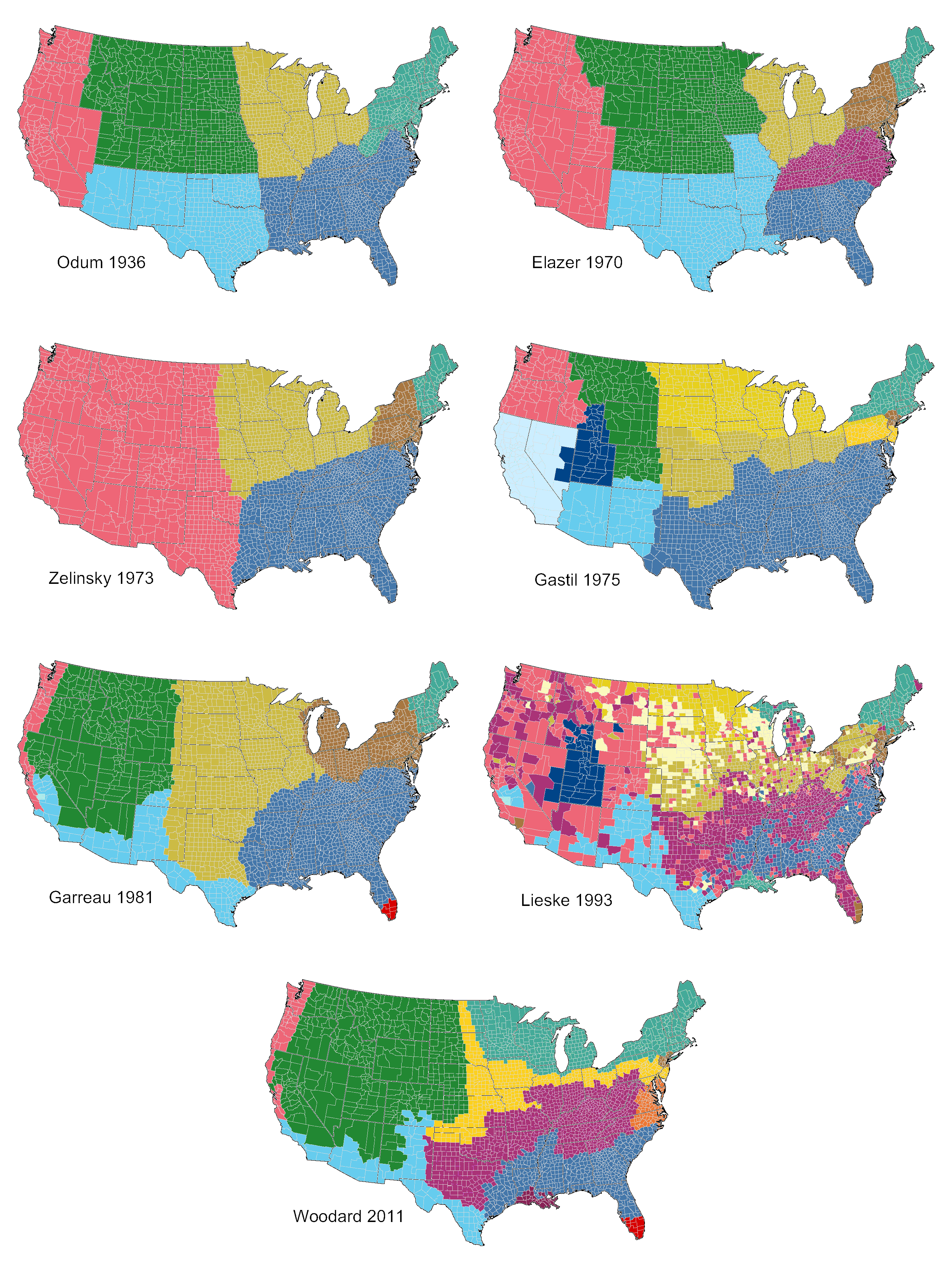}
  \caption{Maps showing the primary American cultural regions as identified in eight
  previous studies at the county level
  \citep{OdumSouthernRegions1936,ElazarCitiesPrairie1970,ZelinskyCulturalGeography1992,GastilCulturalRegions1975,GarreauNineNations1996,LieskeRegionalSubcultures1993,WoodardAmericanNations2012}.}
  \label{fig:litt_cultural_regions}
\end{figure*}

Mapping cultural regions has been a particularly active area of research in the US,
where there has long been debate over the cultural geography of the country, with a wide
range of theories of American cultural regions having been proposed. Seven of the most
prominent theories
\citep{OdumSouthernRegions1936,ElazarCitiesPrairie1970,ZelinskyCulturalGeography1992,GastilCulturalRegions1975,GarreauNineNations1996,LieskeRegionalSubcultures1993,WoodardAmericanNations2012}
are mapped in \cref{fig:litt_cultural_regions}, showing considerable disagreement. For
example, in \citep{ZelinskyCulturalGeography1992} the geographer Wilbur Zelinsky
identified 5 major cultural regions---New England, the Midland, the South, the Middle
West, and the West---based on a synthesis of regional patterns in a wide range of
cultural factors, including ethnicity, religion, economics, and settlement history.
Alternatively, in \citep{GastilCulturalRegions1975} drawing on a similar but more
extensive range of cultural factors, the social scientist Raymond Gastil identified 13
major cultural regions, offering a more complex theory than Zelinsky, including by
dividing Zelinsky's Midland, Middle West, and West regions. The two studies illustrate
two basic limitations with these types of approaches that subjectively synthesize a
range of data to infer cultural regions. First, it is unclear exactly how relevant
cultural factors should be identified. Zelinsky considers fewer factors than Gastil,
which may explain his simpler proposal. Second, it is unclear how these different
factors should be synthesized to produce a single overall map of cultural regions.
Zelinsky places greater emphasis on the importance of initial settlement, which may also
explain his simpler proposal.  

Given the subjectivity underlying these studies, the lack of agreement over the number
and location of American cultural regions (as illustrated in
\cref{fig:litt_cultural_regions}) is not surprising. Only a distinction between the
North and South, reflecting the Union-Confederacy border, and a distinction between the
East and West, reflecting the path of the Rocky Mountains, are common to these most
influential theories of American cultural regions~\citep{OdumSouthernRegions1936,ElazarCitiesPrairie1970,ZelinskyCulturalGeography1992,GastilCulturalRegions1975,GarreauNineNations1996,FischerAlbionSeed1989,LieskeRegionalSubcultures1993,WoodardAmericanNations2012}.
Otherwise, between 4 and 12 primary cultural areas have been mapped, typically including
the Northeast~\citep{OdumSouthernRegions1936,ElazarCitiesPrairie1970,ZelinskyCulturalGeography1992,GastilCulturalRegions1975,GarreauNineNations1996,FischerAlbionSeed1989,LieskeRegionalSubcultures1993}, the South~\citep{OdumSouthernRegions1936,ElazarCitiesPrairie1970,ZelinskyCulturalGeography1992,GastilCulturalRegions1975,GarreauNineNations1996,FischerAlbionSeed1989,LieskeRegionalSubcultures1993,WoodardAmericanNations2012}, the West~\citep{OdumSouthernRegions1936,ElazarCitiesPrairie1970,ZelinskyCulturalGeography1992,GastilCulturalRegions1975,GarreauNineNations1996,WoodardAmericanNations2012},
and the Midwest~\citep{OdumSouthernRegions1936,ElazarCitiesPrairie1970,ZelinskyCulturalGeography1992,GastilCulturalRegions1975,GarreauNineNations1996}.

In large part, the debate over the geography of American cultural regions has been about
which types of cultural factors should be given precedence, and how these factors should be combined. Crucially, these decisions
have generally been left entirely to the judgment of the analyst. Quantitative data from
the census and elections have sometimes been taken into consideration (e.g.~\cite{ZelinskyCulturalGeography1992,GastilCulturalRegions1975,LieskeRegionalSubcultures1993,WoodardAmericanNations2012}), but less often subjected to
statistical analysis (e.g.~\cite{LieskeRegionalSubcultures1993}), while the selection and weighting of these
factors has always been subjective. For example, religion and politics are undoubtedly
important cultural factors, but they can be measured in various ways, and it is unclear
how important they are relatively speaking and whether their importance varies across
the United States.

A basic question is therefore how can we infer general American cultural regions in an
objective way? In particular, how can we both identify a complete or at least
representative range of relevant cultural factors and somehow combine these factors in so as to map American cultural regions?
Defining such regions does not mean that they do not contain internal variation or that they
are separated by hard borders---culture is dynamic and complex and humans are highly
mobile---but that we can find areas where the cultural practice and values of the
people who live within that region are more similar to each other than to those of
people who live outside that region.

The goals of this paper are
therefore to address these issues, by (i) proposing a novel method for discovering
cultural regions by identifying regional patterns in topics of conversation, and by then
(ii) proposing a theory of American cultural regions derived from the application of
this method to a large corpus of geolocated social media data.

Our starting premise is that cultural regions will necessarily be reflected by regional
variation in the topics that people choose to discuss in their everyday lives. If the
cultural geography of the US was broadly homogenous, we would expect topics
of conversation to be largely the same across the country, aside from different uses of
place names and other such relatively superficial and necessarily regionalized vocabulary
items. However, if  people from different regions exhibit distinct and systematic
cultural characteristics—for example, in politics, religion, music, sport, fashion—as
research on American cultural geography has consistently shown, then these patterns of
cultural variation will necessarily manifest themselves as patterns of topical variation
in the language used by people from these regions~\citep{kramsch2014language}. For
example, if hip hop music,  baseball, tattoos, or some other cultural practice is
especially popular in some part of the country, we would expect more discussion on that
topic in large samples of everyday language use originating from that region, including
on social media. Furthermore, previous works show that cultural factors often show related
regional patterns, owing presumably to interrelationships between these different
factors. For example, regional settlement patterns can help explain differences in
ethnicity and religion which can have long term effects on voting patterns.
Consequently, analyzing these regional topical patterns in the aggregate can
be used to infer broader cultural regions. Crucially, there is no need to predefine what
these topical patterns are or how much they matter: the topics themselves and their
relative importance can be inferred through the analysis of everyday language as well.
We therefore introduce an automated method for identifying cultural regions based on the
automated identification of patterns of regional variation in topics of discussion in
very large corpora of geotagged everyday language use. Our method is especially intended
to take advantage of the incredibly large amount of geotagged social media data that
online communication now provides us with for the first time, although our method could
be used to identify cultural regions within any area based on any substantial source of
regionalized everyday language use.

Specifically, to map modern American cultural regions, we identify regional patterns in
the topics that Americans tend to discuss on social media through a quantitative
analysis of ten thousand lexical items in over 3.3 billion geotagged Tweets from across
the US, collected between 2015 and 2021. Large corpora of geotagged Twitter data have
been used frequently in computational sociolinguistics to
identify~\citep{NguyenComputationalSociolinguistics2016}, including to map patterns of
dialect
variation~\citep{GrieveStatisticalMethod2011,EisensteinDiffusionLexical2014,GoncalvesCrowdsourcingDialect2014,HuangUnderstandingRegional2016,GrieveRegionalVariation2016,
Donoso2017,GoncalvesMappingAmericanization2018,AbitbolSocioeconomicDependencies2018,GrieveMappingLexical2019},
while others leveraged methods such as Latent Dirichlet Allocation to identify regional
topical patterns~\citep{KoyluUncoveringGeoSocial2018,FunknerGeographicalTopic2021}.
Despite this wealth of research that has used large corpora of social media to identify
regional patterns in language use, we are aware of no research that has used this type
of information to infer the location of general cultural regions.

Of course, social media or any other form of language can only provide a partial picture
of regional patterns in overall topics of discussion in a region. In general, big data
corpora generated from microblogging platforms certainly present a number of biases:
incomplete demographic representativeness~\citep{MisloveUnderstandingDemographics2011},
particularly for users geotagging their
Tweets~\citep{PavalanathanConfoundsConsequences2015}, non-homogeneous spatio-temporal
distribution~\citep{steiger2015advanced}, or severe topic differences with the offline
world~\citep{diaz2016online}. However, if cultural regions are real and pervasive,
then we should expect these regions to manifest themselves in any large sample of everyday
language that encompass a large proportion of the population, even if the specific
topics of interest vary across these different domains. Furthermore, right now, Twitter
is the only variety of geotagged natural language data available in sufficient
amounts to allow reliable automatic analyses, and is a very popular social
media platform used regularly by millions of people from across the US, mostly in
interactive contexts~\citep{AuxierSocialMedia2021}, serving as a perfect domain to apply
our data-driven approach for automatically mapping cultural regions.

Our main finding is that the modern US can be divided into five primary cultural areas,
each defined by its own topical patterns. We emphasize that this result stems from
a quantitative analysis in contrast to previous proposals based on more or less
informative (qualitative) approaches. Further, beyond the specific number of regions it
is most relevant to note that our method yields the list of words and topics that define
those regions, which highlights the differences in interests, habits and backgrounds
that distinguishes each cultural region from the others. Crucially, by means of a
dynamic analysis we show that cultural regions of the US are relatively stable over the
past few years, offering further evidence that cultural areas are real phenomena that
pervade American society.

The rest of the article is structured as follows. The results of the work are first
introduced by a description of the dataset collection and pre-processing methodology.
Regional variations of words usage observed from this dataset are then explored, before
obtaining the principal dimensions of these variations. The main result of the work, the
cultural regions of the US and the main topics of discussion that define them, is then
presented in detail. The possibility of a variation with time of the results is then
explored. Finally, a discussion of the insights brought by the analysis and also of
where future works could build on it comes to conclude the work.

\section*{Results}

\subsection*{Dataset}

We analyze geotagged Tweets collected through the streaming API of Twitter,
more specifically, using the filtered stream endpoint:
\url{https://developer.twitter.com/en/docs/twitter-api/tweets/filtered-stream}.
This endpoint provides a sample of Tweets in real time matching suitable filters. This allows us to gather 3.3 billion geotagged Tweets from the contiguous US, posted from January $1^\text{st}$, 2015 to December $31^\text{st}$, 2021. Importantly, we discard
users tweeting at an inhuman rate, which we define to be any rate superior to 10 Tweets
per hour over one's whole tweeting span. We also discard users tweeting from any
platform that is not a Twitter mobile application or their website. In our
dataset, we thus retain 17 million users. We strip Tweets of any link, hashtag or user
mention, and only keep those which still have more than 4 words after this filter. Hashtags were discarded out of precaution: some of them may be content words, but they may also be related to short-lived trends for instance. As we found that the content of hashtags accounted for less than $\SI{5}{\percent}$ of the content of the Tweets we collected, they can anyway be safely discarded. We
subsequently use the Chromium Compact Language Detector (CLD) \citep{Al-RfouPythonBindings2014} to
eliminate Tweets written in a language other than English. To attach a geolocation to Tweets, they are geotagged with either the precise GPS coordinates of the device of the user, or ``places'', which can be an administrative region, a city or a place of interest. Then, as these geotags may be places of
the size of a state, we also remove Tweets with a geotag that did not allow for
reliable assignment to our unit areas, which are the US counties and county equivalents
($\SI{3108}{}$ in total). Certainly, counties vary in both size and population but most of them
form a useful division sufficiently large to show a sizeable amount of Tweets and
sufficiently small to allow for a careful delimitation of cultural areas (states would
be too big units whereas towns would be too small).

From the remaining Tweets, we extract and count the tokens in their text, and assign
them to counties. Counties that accumulate fewer than \SI{50000}{} tokens are not taken
into account, leaving us with $N_c = \SI{2576}{}$ counties which define our sub-corpora.
We thus keep $\SI{83}{\percent}$ of the total number of counties. After this filtering,
the full dataset contains $9.1$ billion tokens (see Table S1 for a summary description
of the dataset).
We subsequently convert the remaining word forms to lowercase and aggregate the token counts on
these forms. We then remove all function words (like \textit{the, and})
and interjections (like \textit{um, oh}) (see Data availability for access to the full
list of exclusions), and consider the \SI{10000}{} most common remaining word forms.
Note that this list of word forms emerges from the data, and is not imposed by any
previous topical or dialect classification.

\begin{figure*}
  \centering
  \includegraphics[width=0.95\textwidth]{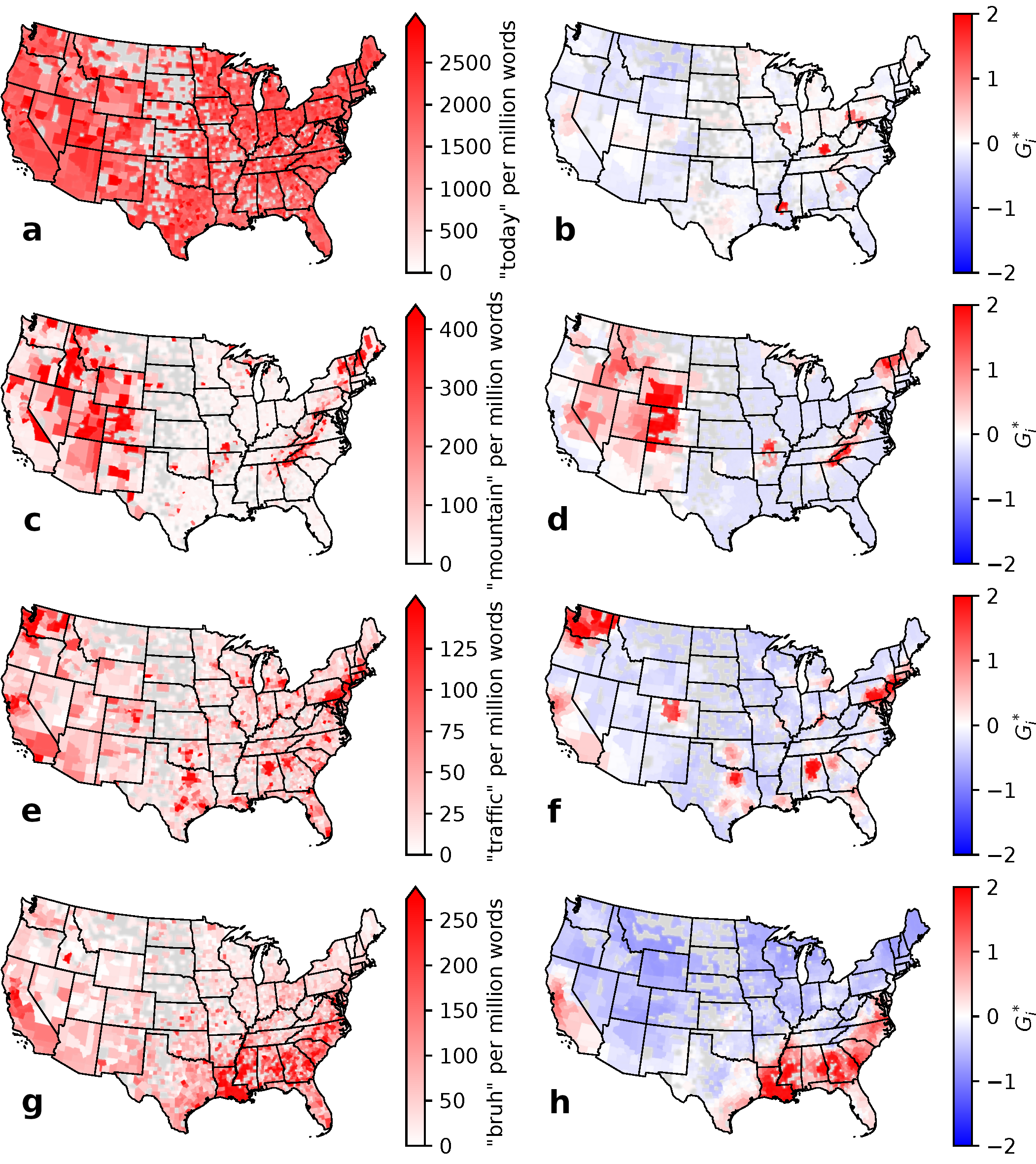}
  \caption{Maps showing the \textbf{a-c-e-g} relative frequency and \textbf{b-d-f-h} Getis-Ord  $G_i^*$ z-score
  for the words \textit{today}, \textit{mountain}, \textit{traffic} and \textit{bruh}, respectively.
  One can note how the latter metric enables to reveal word usage hotspots, smoothing out the raw noisy signal from the data.}
  \label{fig:example_words_maps}
\end{figure*}

\subsection*{Measuring regional variation}

We then measure and map the relative frequencies $f_{c, w}$ for every word $w$ in every county $c$. We illustrate our raw results by plotting
in \cref{fig:example_words_maps} the relative frequency in each county of four
representative words: \textbf{a} \textit{today}, \textbf{c} \textit{mountain}, \textbf{e} \textit{traffic}
and \textbf{g} \textit{bruh} (cells that appear greyed out do not reach a minimum number of
Tweets as explained in the paragraph above). In the first case, \textit{today} appears at relatively stable rates
in most of the counties, as expected. Alternatively, \textit{mountain} is a
regionally-dependent word as clearly seen. The item \textit{traffic} appears more
frequently in urban areas. Finally, \textit{bruh} is an African-American English variant
that appears to be especially common in southern counties, where there are large African-American populations.

A word of caution is now required. A relative frequency map alone is not able to fully
reveal regional variations due to the wide range of different factors besides regional variation that affect word use and add noise to the signal. To extract the underlying regional signal from each
word map, we conduct a multivariate spatial analysis \citep{GrieveStatisticalMethod2011,GrieveRegionalVariation2016} of the
relative frequencies of our \SI{10000}{} word forms. In order to identify geographical
hotspots in the usage of each word (\cref{fig:example_words_maps}), we compute
Getis-Ord's z-scores ($G_i^*$ \citep{OrdLocalSpatial1995}) for each county $c$ and word $w$, which
are defined as:
\begin{equation}
\label{eq:Gi_star}
  G_{c, w}^* = \frac{
      \sum_{c'} W_{c, c'} (f_{c', w} - \bar{f_w})
    }{
      \sigma_w \sqrt{\frac{
        N_c \sum_{c'} W_{c, c'}^2
          - \left( \sum_{c'} W_{c, c'} \right)^2
        }{
          N_c - 1
        }
      }
    },
\end{equation}
with $\bar{f_w}$ the average frequency of $w$ over the whole dataset, $\sigma_w$ the
standard deviation in $w$'s frequencies, and $W_{c, c'}$ are the elements of a proximity
matrix, which we take as equal to $1$ if $c' = c$ or $c'$ belonging to $c$'s 10
nearest neighbors, and equal to $0$ otherwise.

The metric given by \cref{eq:Gi_star} ultimately diminishes spurious data variation
and smooths spatial patterns,
allowing us to
discern a regional pattern in a word's usage.
In \cref{fig:example_words_maps}\textbf{b}, \textbf{d}, \textbf{f} and \textbf{h} we show,
respectively, the $G_i^*$ z-scores for the previous
words \textit{today}, \textit{mountain}, \textit{traffic} and \textit{bruh}.
White, light blue or light red counties do not depart significantly
from an average utilization, whereas a bright red or blue respectively 
mean that the word is relatively frequently or infrequently used in that region.
Since \textit{today} is a rather generic word, we do not find any
strong regional pattern, whereas the others do. The usage hotspots of \textit{mountain} display the main mountain ranges of the country.
While the map for \textit{traffic} is correlated with large urban areas (and can be interpreted as a topical word), the dialect word \textit{bruh} seems to be significantly more used
in counties pertaining to the Deep South.
We see here that different attributes that define a culture
(interests, behavior, dialect) are captured within our scheme
and, notably, are treated on equal footing.

\subsection*{Obtaining the principal dimensions of regional variation}

The $G_i^*$ distributions for all \SI{10000}{} top words by usage thus hold valuable information. However, a considerable part of this information can be analyzed more efficiently, since some words may belong to the same semantic field (\textit{mountain} and \textit{peak}) or characterize the same particular dialect (\textit{bruh} and \textit{aight}). Furthermore, a few variations may simply be uninformative noise, intrinsic to real individuals' behaviors, but also potentially resulting from an imperfect filtering of Twitter data, as aforementioned.
The most important dimensions of regional lexical variation are then found by
subjecting the hotspot maps for the complete set of words to a principal component
analysis \citep{LieskeRegionalSubcultures1993,WoldPrincipalComponent1987}. 
Another possible approach would have consisted in performing topic modelling, for instance by ways of a Latent Dirichlet Allocation on the word frequency matrix, to then infer a distribution of topics for every county. It is however more computationally intensive, and poses the questions of the selection of the number of topics, their interpretability and their internal coherence \citep{ArunFindingNatural2010,HasanNormalizedApproach2021}.
In a case like ours where documents are so large (aggregating all Tweets in a county), it is far from obvious to select a number of topics such that there is little overlap between them, and to know that these topics are actually representative of the dataset as a whole. This is much more clear when selecting components in PCA, as we show below.

\begin{figure*}
  \centering
  \includegraphics[width=1\textwidth]{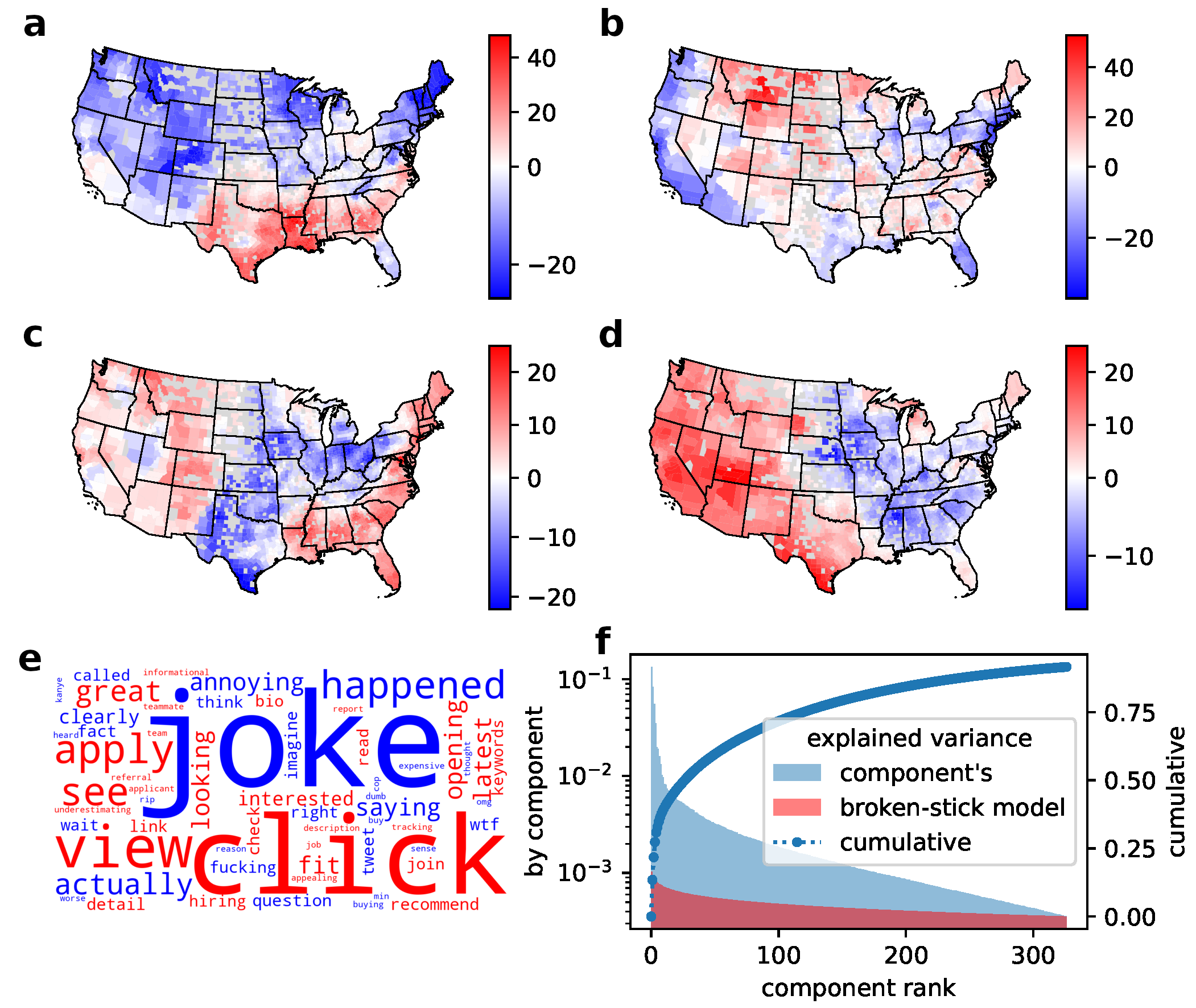}
  \caption{Result of the principal component analysis carried out on our whole dataset. \textbf{a-d} Four maps show the projection of the data along the first four components, highlighting regional lexical variations. Note that the scale on the divergent color scales are not symmetrical around zero in order to utilize the full range of colors of the color map. \textbf{e} Word cloud showing the words with the strongest positive (red) and negative (blue) loadings for the second component, with each word's font size depending on its loading's absolute value. \textbf{f} Explained variance of the principal components compared to the broken-stick model on a logarithmic scale, which shows how the number of components to keep is selected at the first intersection of the two curves. The cumulative proportion of the variance explained by the components is also plotted, showing that our dimension reduction explains around 92\% of the observed variance.
  The first four components shown in panels \textbf{a-d} capture alone 31\% of the variance.}
  \label{fig:whole_data_decomp}
\end{figure*}

From the $N_w = \SI{10000}{}$ dimensions of our dataset, we
thus project to a principal component (PC) space of $N_{\text{PC}} = 326$
dimensions. It turns out that these 326 components explain 92~\% of the observed variance (see \cref{fig:whole_data_decomp}\textbf{f}).
We do not set this number of components arbitrarily, by choosing one directly
or by setting a percentage of variance we wish to explain using these components.
Instead, we use the broken-stick rule to fix the number of components
\citep{FrontierEtudeDecroissance1976,JacksonStoppingRules1993}. This heuristic compares the decrease of the variance
explained by each successive component to the one expected from a random partition of the whole
variance in $N_w$ parts. Components, sorted by decreasing explained variance, are kept
until they do not explain more variance than their corresponding random part would. With
this method, we do not make any assumption about the amount of variance in our data that
is simply due to random fluctuations.

We show the projected data
along the first four PCs in \cref{fig:whole_data_decomp}\textbf{a-d},
which displays a neat visualization of the spatial patterns.
The map for each dimension shows two opposing regions (red and blue)
which can be linked to their characteristic words, the ones with the highest (positive, in red) and lowest (negative, in blue) loading.
For an illustration, in \cref{fig:whole_data_decomp}\textbf{e} we show in a word cloud the most characteristic words for each of the two regions in \cref{fig:whole_data_decomp}\textbf{b}, which corresponds to the second component. In Figs. S2 and S3 we plot,
respectively, the projected data for the proximity matrix $W_{c, c'}$ of \cref{eq:Gi_star} defined based on 5 and 15 nearest neighbors. The results show that the
components are not significantly altered by a slight change of the proximity matrix.
Figure S4 shows the results when $W_{c, c'}$ is alternatively defined in terms of a fixed distance. In
this case, the modifications are stronger because the size of the counties is not
uniform. This demonstrates that one should take proper neighbor couplings when dealing
with heterogeneous geographical units.

\subsection*{Inferring cultural regions}

\begin{figure*}
\centering
  \includegraphics[width=0.85\textwidth]{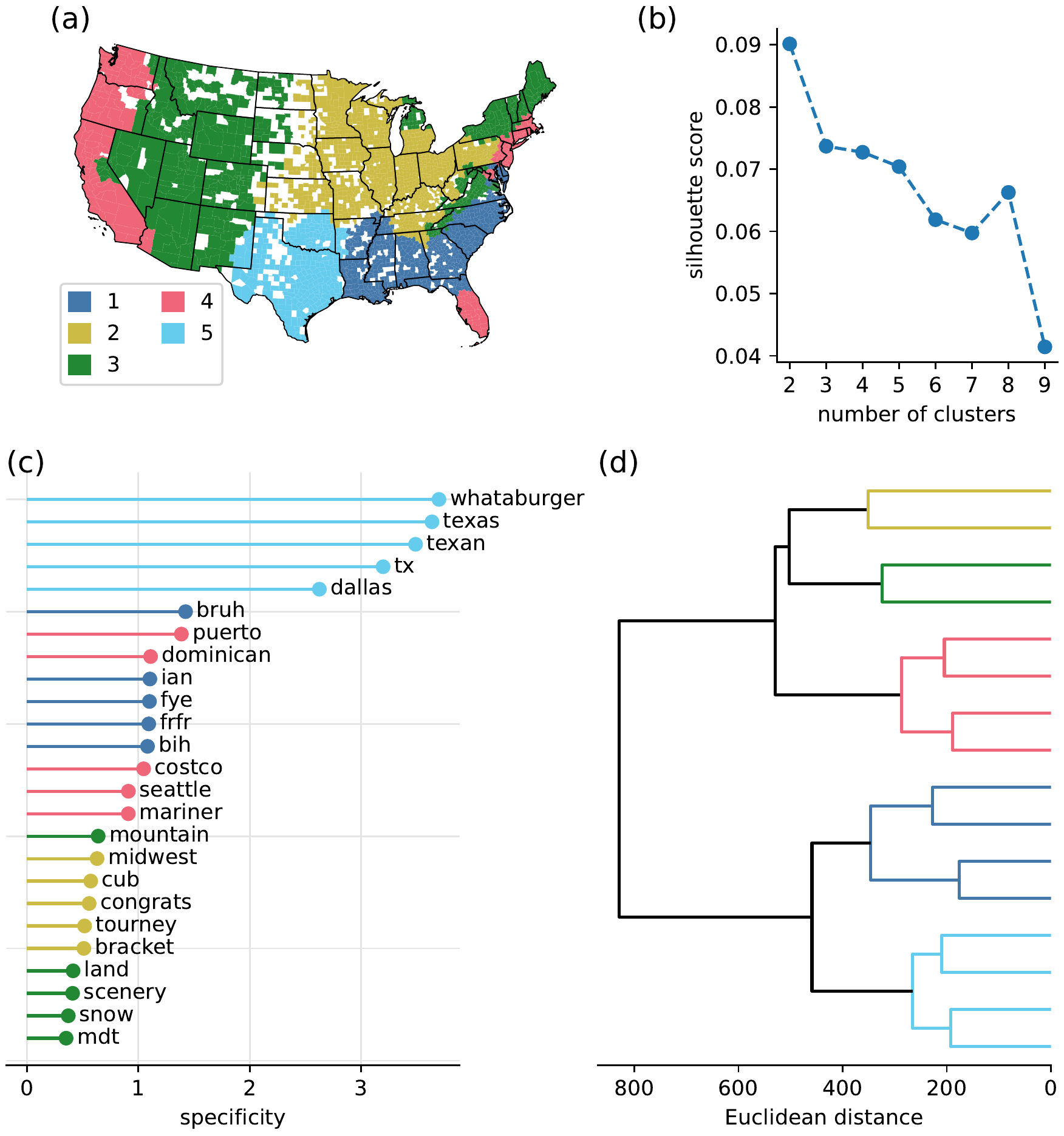}
  \caption{Cultural regions obtained from our whole dataset. \textbf{a} Map of the five clusters obtained through hierarchical clustering, selected from a high value of \textbf{b} the mean Silhouette score. A significant drop of the Silhouette score after 5 levels indicates that further splitting counties in more region does not yield coherent cultural regions. \textbf{c} Five most specific words for the five clusters shown in \textbf{a}, along with their specificity values. \textbf{d} The dendrogram allows seeing which clusters are first joined if going to a higher level of clustering, and thus which ones are closer together. It clearly shows that the strongest division is the one between the North and the Southeast (excluding Florida) with further splittings as the cluster distance increases.}
  \label{fig:whole_data_clust}
\end{figure*}

We are now in a position to generate a single overall taxonomy of American cultural
regions by clustering together counties with similar lexical signature. To do so, we
subject the previous PC maps to a hierarchical clustering, using the Euclidean distance
and the Ward variance minimization algorithm \citep{EverittClusterAnalysis2011}. This is how we define
the cultural regions from our corpus, as depicted in \cref{fig:whole_data_clust}.
From the dendrogram and the evolution of the average silhouette score for different levels of
clustering, we select a meaningful number of clusters $n_{\text{clusters}}$
\citep{RousseeuwSilhouettesGraphical1987}. The hierarchical nature of the clustering is useful to see
how regions are grouped together at different levels of clustering, indicating which
regions are closer together. Importantly, applying
hierarchical clustering to the principal dimensions of variation of the data obtained through PCA allows us to focus on the main regional patterns of variation. Applying the algorithm directly to the
$\SI{10000}{}$-word distance matrix would yield highly noisy results.

We plot the main divisions in \cref{fig:whole_data_clust}\textbf{a}. This is 
the main result of our paper. In the map, we present the division into five clusters since
it is one of the two best options as characterized by the Silhouette score analysis in
\cref{fig:whole_data_clust}\textbf{b}, and at a clear-cut on the dendrogram in
\cref{fig:whole_data_clust}\textbf{d}. The optimal choices correspond to the two significant drops in the
score: the first (second) corresponds to a cluster number equal to 2 (5). 

Indeed, the dendrogram in \cref{fig:whole_data_clust}\textbf{d} shows that the counties can be 
initially classified into two large-scale subgroups representing a North vs South divide. The North is then further fragmented into
the clusters 2, 3 and 4 shown in \cref{fig:whole_data_clust}\textbf{a},
whereas the South group splits into the clusters 1 and 5.
For the most part, our
map in \cref{fig:whole_data_clust}\textbf{a} is consistent with standard theories of American cultural regions, with all five of
our regions finding analogs in existing systems. Yet,
taken as a whole our clusters do not match any previous system and reveal 
non-contiguous culture regions such as the clusters 3 and 4.
Moreover, in contrast to previous proposals our results have the
advantage of being data-driven, based on variation in the topics people care to discuss as opposed to factors selected by hand by the researcher
(and consequently subjected to many more, uncontrolled biases than our Twitter data).

Further, to be able to better interpret the obtained regions, it is
insightful to know which words characterize each cluster the most. To infer them, we
start by taking the center of each cluster in words-$G_i^*$-space. Hence, for
each cluster we take the average $G_i^*$ score over its counties for all words. From
these $n_{\text{clusters}}$ vectors of $N_w$ elements, we calculate the minimum absolute
difference between each cluster center's word's score and the ones of all other
clusters, i.e., we take the distance to the closest cluster's center along
the word's dimension. More formally, we define the specificity $S_{C, w}$ of word $w$ for cluster $C$ as:
\begin{equation}
\label{eq:spec}
  S_{C, w} = \min_{C' \; \in \; \mathcal{C} \setminus C} \left(
  \frac{1}{N_C} \sum_{c \; \in \; C} G_{c, w}^*
    - \frac{1}{N_{C'}} \sum_{c \; \in \; C'} G_{c, w}^*
  \right)^2 ,
\end{equation}
where $\mathcal{C}$ denotes the set of clusters, $N_C$ the number of counties belonging to cluster $C$, and $G_{c, w}^*$ the $G_i^*$ score of word $w$ in county
$c$. For each cluster $C$, we thus define the most characteristic words as the ones with
highest $S_{C, w}$ values.
In the case of the division in five clusters, the top 5 most characteristic words per cluster are shown in \cref{fig:whole_data_clust}\textbf{c}, according to the specificity metric defined in \cref{eq:spec}.
In all cases, the five cultural regions are linked to clear and distinct topical patterns (see the Supplementary Information for a more exhaustive list).
We stress that these characteristic words 
are automatically identified based on the quantitative analysis
presented above. Notably, for each cluster we see three basic types of lexical patterns.

First, we see words associated directly with those locations, most commonly
the names of cities, states, and sports teams. This is basic evidence that the method 
works: we would expect these words to be associated with the cultural regions that
 contain them. However, these results also reflect how often people from different 
cities and states refer to each other. For example, the fifth cluster which is centered 
on Texas also includes Oklahoma, which contributes various place names to the
list of words most strongly associated with this region. This means not only that
people in Texas and Oklahoma talk more about place names in their own states, 
as would be expected, but that they talk more about place names in each other's
states. This is one type of regional topical patterns that our approach draws to
identify cultural regions.

Second, we observe words connected with non-regional topics, which 
nonetheless show regional differences. In this case, our approach can be seen as
discovering topical patterns and by extension cultural patterns that distinguish 
between different regions of the US. For example, cluster 2 is strongly associated with the discussion of a range of American sports, as well as the names of the states that fall within this region. 
Although we would expect that a region centered around the Midwest would be
associated with the names for Midwestern states, their preoccupation with the discussion of
sports on Twitter is not so easy to predict. 

Third, we find words that are dialect items, i.e., alternative ways of referring
to a given concept. This type pattern is especially apparent for
cluster 1, which aligns closely with the region of African American
population density and is therefore associated with numerous lexical items from 
African-American English (e.g. \textit{bruh, lawd, turnt}). Although dialectologists do not
usually focus on the frequencies of individual words, this results is to be expected: 
dialect regions, which can be seen as a type of cultural region, have been found
to generally align with broader cultural regions~\citep{GrieveRegionalVariation2016}.

We can now examine each of the five cultural regions we have identified in turn and consider what the words that are mostly strongly associated with each tell us about the culture of that region, as well as the factors that drive cultural variation in the US more generally. 

The first cluster [blue in \cref{fig:whole_data_clust}\textbf{a}], which identifies a southeastern region, largely reflects African American culture, as can be predicted based on the close correlation between our map and the distribution of counties with relatively large African American populations (see Fig.~S1). Most notably, Tweets from the South are more likely to contain words related to African American culture, including, for example, cuisine (e.g. \textit{grits, cookout}), fashion (e.g. \textit{braids, dreads}), and music (e.g. \textit{rappers, rapping}). As noted above, this cluster is also strongly characterized by many vocabulary items associated with African American English, especially for referring to people (e.g. \textit{bruh, dawg}), as well as many acronyms (e.g. \textit{frfr, stg}). Place names associated most strongly with this cluster primarily include southern states (e.g. \textit{Georgia, Carolina}), despite the fact that, in general, references to place names are relatively rare compared to other clusters.

The second cluster [yellow in \cref{fig:whole_data_clust}\textbf{a}] has its core in the Midwest and is clearly characterized by more frequent references to sports. American team sports especially stand out, with 40 words of the top 50 most strongly associated with this cluster being directly linked to this topic. In particular, these are words associated with basketball (e.g. \textit{basketball, rebound}) and baseball (e.g. \textit{baseball, innings}), although football, wrestling, and cheering are also referenced, as well as various more generic sporting terms (e.g. \textit{teams, tourney}). Similarly, many place names are associated with local sports teams (e.g. \textit{Cubs, Chiefs}), although various state names are also strongly associated with this cluster (e.g. \textit{Ohio, Illinois}), as well as the word \textit{Midwest} itself.  A smaller number of lexical items are also associated with school (e.g. \textit{locker, choir}). Overall, this cluster therefore shows that sports is a central part of this region. 

The third cluster [green in \cref{fig:whole_data_clust}\textbf{a}] can be identified with a discontinuous region that mostly aligns with rural areas of the US, as well as areas that focus on outdoor activities, especially in mountainous regions (e.g., the Rocky or Appalachian Mountains). This cluster is relatively hard to interpret topically, in part because, unlike the other regions, it is characterized by the relative infrequent use of a number of words. In terms of words that are relatively common in this region, the clearest pattern is a relatively large number of words associated with nature (e.g. \textit{mountains, tree}), weather (e.g. \textit{snow, seasonal}), and outdoor activities (e.g. \textit{adventures, trail}). Clearly, people in this region tend to focus more of their natural surroundings. In addition, there are a number of words related to work (e.g. \textit{hiring, jobs}), as well as a numerous place names (e.g. \textit{Colorado, Montana}) that are strongly associated with this region. In terms of words that are uncommon within the cluster, there exist many verbs, especially verbs associated with human actions like communication (e.g. \textit{said, told}), thought (e.g. \textit{understand, confused}), and physical actions (e.g. \textit{put, hit}), which implies overall less focus on the individual. This region is also associated with relatively infrequent use of a wide range of negative words (e.g. \textit{wrong, bad}), which largely hints at a more positive outlook.  

The fourth cluster [red in \cref{fig:whole_data_clust}\textbf{a}] also identifies a discontinuous region that primarily encompasses large urban areas in the coasts (Northeast and West). Unsurprisingly, this region is characterized by a wide range of words associated with more urban life (e.g. \textit{homeless, traffic}), especially terms related to different nationalities and immigration (e.g. \textit{Latino, Asian}). We also find a relatively large number of place names (e.g. \textit{California, NYC}). Strikingly, this cluster is associated with a very large number of words with negative connotations, including relating to violence (e.g. \textit{violence, attack}), danger (e.g. \textit{dangerous, crime}), cursing (e.g. \textit{asshole, fucking}), political unrest (e.g. \textit{protests, indicted}), racism (e.g. \textit{Nazi, supremacist}), and general negative adjectives (e.g. \textit{disgusting, abusive}). Quite generally, people from this cluster are more likely to discuss negative topics than other parts of the US, at least on social media. Taken together, the third and fourth clusters suggest an opposition in the culture of more rural and urban areas in the US, which appear to engage in more positive and negative discourse respectively~\citep{vanderbeck2003young}. 

Finally, the fifth cluster [cyan in \cref{fig:whole_data_clust}\textbf{a}], which is centered around South Central States, especially Texas and Oklahoma, is characterized by frequent reference to place names, relative to the other clusters, especially in these two states, as has already been noted. For example, the first five most strongly associated words are \textit{Whataburger} (a fast food chain from Texas), followed by \textit{Texas, TX, Texan, and Dallas}. This not only shows that people in this region tend to discuss place more on Twitter, but implies that this cultural region is characterized by a relatively high amount of local pride. Correspondingly, this region is also associated with a relatively large number of dialect terms, both of Anglo (e.g. \textit{yalls, fixing}) and Hispanic (e.g. \textit{queso, taco}) origins, reflecting the diverse makeup of this region.

The analysis yielding \cref{fig:whole_data_clust} was repeated, adding a stemming step at the very beginning of our pipeline. We obtain a very similar result, shown in Fig.~S5, indicating little sensitivity of our results to stemming.

Given the lack of consensus in previous research, our results can help resolve long-standing debates relating to the distribution of American cultural regions. 
We find that the division between the Southeast and the rest of the US is the strongest.
This result attests to the importance of the cultural divide between White and Black
America and between the North and the South. Although all previous major theories of American cultural regions has identified a distinction between the North and the South, our southern region is especially similar to relatively recent theories, which identify a southern region that closely aligns with the part of the south with an especially high proportion of African Americans \citep{LieskeRegionalSubcultures1993,WoodardAmericanNations2012}. Another key finding that emerged from our analysis is a broad opposition between coastal
and internal areas, which has not previously been
identified as important sources of distinction of American cultural regions
\citep{OdumSouthernRegions1936,ElazarCitiesPrairie1970,ZelinskyCulturalGeography1992,GastilCulturalRegions1975,GarreauNineNations1996,FischerAlbionSeed1989,LieskeRegionalSubcultures1993,WoodardAmericanNations2012}
but reflects a modern political trend of undeniable
significance \citep{GelmanRedState2009} that is currently reconfiguring the nation. The discontinuous nature of these regions, which is not required by our definition of a cultural region, is also notable. It demonstrates how patterns in American culture can be distributed across very wide areas, reflecting complex patterns in physical and human geography, and the underlying complexity and dynamic nature of American society. This result is broadly in line with other recent theories of American cultural regions which have also identified discontinuous cultural regions \citep{LieskeRegionalSubcultures1993,WoodardAmericanNations2012}.

Our analysis is further useful for understanding the relationship between these regions. It
divides the South into two regions, splitting Texas off from the rest of the Southeast,
and splits the Midwest off from the rest of the North, divided into discontinuous
countryside/coastal regions, rather than contiguous cultural regions.
However, on the question of the number of primary American cultural regions, we can only
safely say that with our data and methodology, at least 5 distinct regions can be
discerned. We do not see it here, but we still cannot discard recent theories that claim that
America is fundamentally far more culturally fragmented
\citep{GarreauNineNations1996,LieskeRegionalSubcultures1993,WoodardAmericanNations2012}.

\subsection*{Temporal aspect of the results}
Given the success of our analysis, it would be interesting to see how the cultural regions found in
\cref{fig:whole_data_clust} change with time, as has been done in other research analyzing diachronic corpora
\citep{BochkarevAverageWord2015,BentleyBooksAverage2014,KarjusQuantifyingDynamics2020,MomeniModelingEvolution2018,AlshaabiStorywranglerMassive2021}.
Although we would not expect significant changes due to the short timescale imposed by
our Twitter dataset, we can still carry out a diachronic study to validate the very
existence and meaningfulness of the cultural regions. To do so, we split our corpus into
three datasets corresponding to different year ranges: 2015-2016, 2017-2018 and
2019-2021. These periods have a similar amount of tokens and can be then subjected to
comparison (see Supplementary Table 1). We show their maps in
\cref{fig:evol_comp1_dists}\textbf{a-c}. We obtain similar patterns, despite the variety of
topics and forms employed on Twitter along the years and the heuristic nature of the
clustering method that introduces a small amount of noise in the results. The
North-South division is stable over time with small variations that can be due to either
fluctuations or incipient structural changes. The latter cannot be conclusive due to the
short time period considered in this work.

\begin{figure*}
  \centering
  \includegraphics[width=1\textwidth]{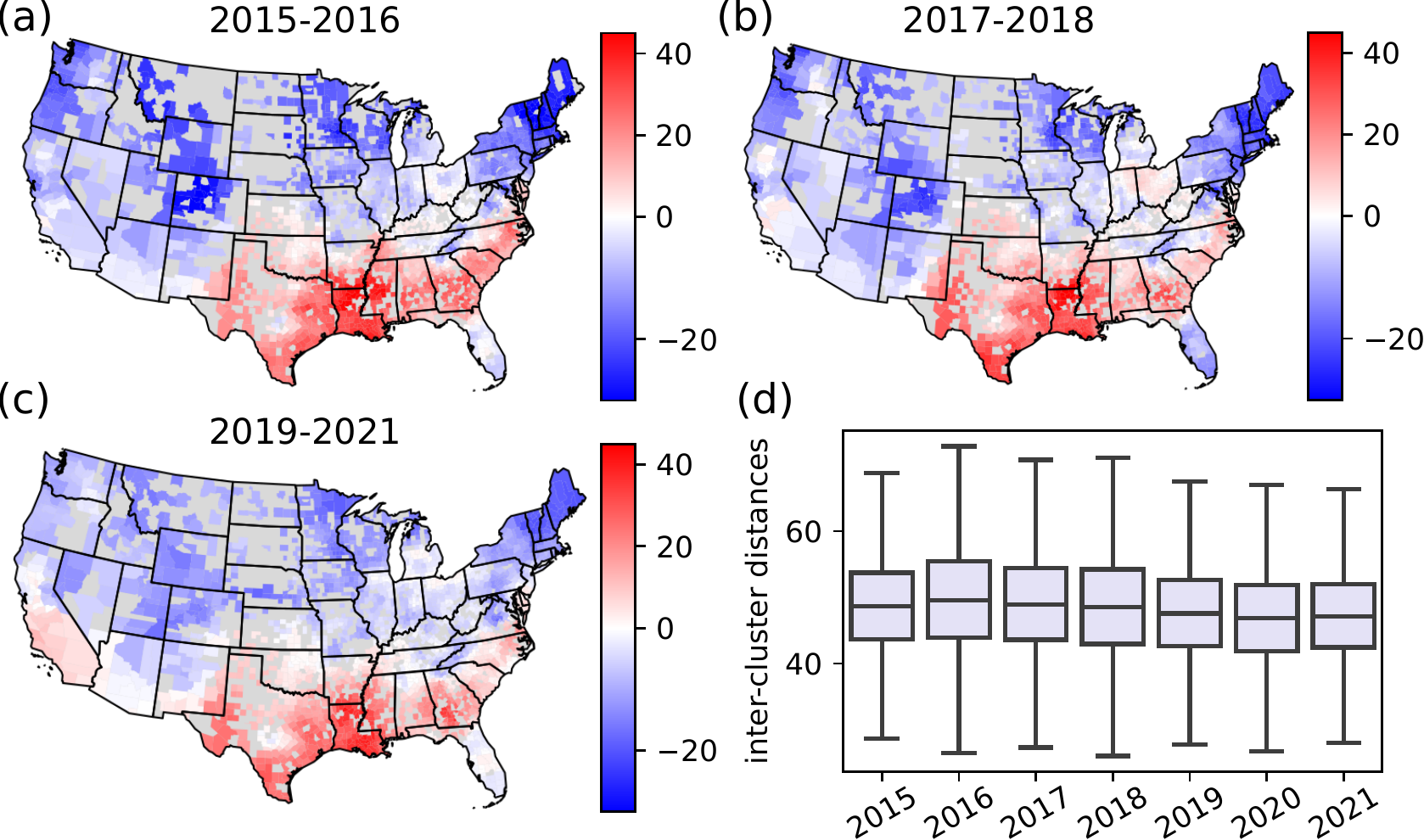}
  \caption{Effect of temporal segmentation for the data on the obtained divisions. \textbf{a-c} Maps of the data projected along the first PC obtained for the years 2015-2016, 2017-2018 and 2019-2021, respectively. Apart from slight variations in California and Florida, the first component translates the same division between the Southeast and the rest of the US. Note that the scale on the divergent color scales are not symmetrical around zero in order to utilize the full range of colors of the color map. \textbf{d} Evolution of the distributions of inter-cluster distances along the years spanned by our dataset. The box plots show the median, first and third quartile, and the boundaries of the whiskers are within the 1.5 interquartile range value. We use the cluster assignment obtained with the whole dataset and measure the Euclidean distance in $G_i^*$ space between counties belonging to different counties. The distribution is thus shown to vary little from year to year,  which demonstrates the stability of the two-way division we found.}
  \label{fig:evol_comp1_dists}
\end{figure*}

Next, we take the hierarchical clustering in \cref{fig:whole_data_clust}\textbf{d} and select the county-to-cluster assignment corresponding to the highest level of the hierarchy. This is represented by the two-way division between North and South. For each year in our dataset, we then measure the pairwise distances between counties belonging to both clusters. The distances are calculated as Euclidean distances between rows of the matrix $G_{c, w}^*$ (see \cref{eq:Gi_star}). We thus obtain the evolution with time of the inter-cluster distances distribution as shown in \cref{fig:evol_comp1_dists}\textbf{d}. The box plots demonstrate that (i) the median distance is roughly constant over the years, and (ii) the distance distribution shows little variation. Both findings suggest that the detected cultural regions are no artifact of the method, but a genuine data structure that exists within our corpus.

\section*{Discussion}

Overall, our analysis has therefore identified regional patterns of lexical variation of
clear cultural importance. Furthermore, the themes associated with each of these
patterns provide a new perspective on American cultural geography. For example, although
our analysis has confirmed that factors such as ethnicity and religion are important for
defining American cultural regions, we found substantial variation in the relevance of
these factors across the US. Our analysis has also identified other subtler cultural
patterns---such as a focus on social interaction, the outdoors, family, and leisure---which have been overlooked in previous research, in part because they cannot be
easily studied through the analysis of traditional sources of secondary data. Our method
has therefore not only allowed us to map cultural regions, but it has also allowed us to
identify cultural factors that are important for defining these regions, at least in this communicative context, providing a foundation for a more
complete picture of the American cultural landscape.

Clearly, our study has only analyzed one genre of
American English. The specific topical patterns on Twitter would
not be exactly replicated in other genres, especially given the communicative purpose and user base associated with microblogging platforms. Nevertheless, assuming that American cultural regions
are important and pervasive forces, similar regional
patterns should be reflected across all genres. This issue could be further clarified when more richly annotated natural language data becomes available in a near future.
Our methods, however, will remain valid for any such dataset. Crucially, we expect that our main idea of inferring cultural regions and the topics defining them from people's speech will be applicable to any big data resource with linguistic value.

\section*{Code availability}
The data processing and plotting of results were carried out in Python with the help of open-source libraries. All code used for this work is hosted on GitHub \citep{LoufThomasWordsuse2023}.

\section*{Data availability}
All aggregated data generated by our Twitter data analyses as well as our list of
excluded words are available for download from a figshare repository \citep{LoufThomasWordCounts2023}. County and state boundary shapefiles from the US
census of 2018 that we used to draw our maps are freely available for download at
\url{https://www.census.gov/geographies/mapping-files/2018/geo/carto-boundary-file.html}.

\section*{Acknowledgments}
This work was partially supported by the Spanish State Research Agency
(MCIN/AEI/10.13039/501100011033) and FEDER (UE) under project APASOS (PID2021-122256NB-C21) and the Mar{\'\i}a de Maeztu project CEX2021-001164-M, by
the Government of the Balearic Islands CAIB fund ITS2017-006 under project PDR2020/51, and by the Arts and
Humanities Research Council (UK), the Economic and Social Research Council (UK), Jisc
(UK) (Jisc grant reference number 3154), and the Institute of Museum and Library
Services (US), as part of the Digging into Data Challenge (Round 3).

\section*{Author contributions}
All authors designed research and performed the data analyses. T. L. built the corpus
and wrote the code that produced the results. J. G and T. L. drafted the initial
manuscript, and all authors were involved in subsequent revisions. J. J. R., D. S. and
J. G. acquired funding.

\section*{Competing interests}
The authors declare no competing interests.

\section*{Ethical approval}
This article does not contain any studies with human participants performed by any of the authors.

\section*{Informed consent}
This article does not contain any studies with human participants performed by any of the authors.

\bibliography{biblio.bib}

\begin{thebibliography}{46}%
\makeatletter
\providecommand \@ifxundefined [1]{%
 \@ifx{#1\undefined}
}%
\providecommand \@ifnum [1]{%
 \ifnum #1\expandafter \@firstoftwo
 \else \expandafter \@secondoftwo
 \fi
}%
\providecommand \@ifx [1]{%
 \ifx #1\expandafter \@firstoftwo
 \else \expandafter \@secondoftwo
 \fi
}%
\providecommand \natexlab [1]{#1}%
\providecommand \enquote  [1]{``#1''}%
\providecommand \bibnamefont  [1]{#1}%
\providecommand \bibfnamefont [1]{#1}%
\providecommand \citenamefont [1]{#1}%
\providecommand \href@noop [0]{\@secondoftwo}%
\providecommand \href [0]{\begingroup \@sanitize@url \@href}%
\providecommand \@href[1]{\@@startlink{#1}\@@href}%
\providecommand \@@href[1]{\endgroup#1\@@endlink}%
\providecommand \@sanitize@url [0]{\catcode `\\12\catcode `\$12\catcode `\&12\catcode `\#12\catcode `\^12\catcode `\_12\catcode `\%12\relax}%
\providecommand \@@startlink[1]{}%
\providecommand \@@endlink[0]{}%
\providecommand \url  [0]{\begingroup\@sanitize@url \@url }%
\providecommand \@url [1]{\endgroup\@href {#1}{\urlprefix }}%
\providecommand \urlprefix  [0]{URL }%
\providecommand \Eprint [0]{\href }%
\providecommand \doibase [0]{https://doi.org/}%
\providecommand \selectlanguage [0]{\@gobble}%
\providecommand \bibinfo  [0]{\@secondoftwo}%
\providecommand \bibfield  [0]{\@secondoftwo}%
\providecommand \translation [1]{[#1]}%
\providecommand \BibitemOpen [0]{}%
\providecommand \bibitemStop [0]{}%
\providecommand \bibitemNoStop [0]{.\EOS\space}%
\providecommand \EOS [0]{\spacefactor3000\relax}%
\providecommand \BibitemShut  [1]{\csname bibitem#1\endcsname}%
\let\auto@bib@innerbib\@empty
\bibitem [{\citenamefont {Broek}\ \emph {et~al.}(1973)\citenamefont {Broek}, \citenamefont {Webb},\ and\ \citenamefont {Hsu}}]{broek1973geography}%
  \BibitemOpen
  \bibfield  {author} {\bibinfo {author} {\bibfnamefont {J.~O.~M.}\ \bibnamefont {Broek}}, \bibinfo {author} {\bibfnamefont {J.~W.}\ \bibnamefont {Webb}},\ and\ \bibinfo {author} {\bibfnamefont {M.-L.}\ \bibnamefont {Hsu}},\ }\href@noop {} {\emph {\bibinfo {title} {A Geography of Mankind}}}\ (\bibinfo  {publisher} {{McGraw-Hill New York}},\ \bibinfo {year} {1973})\BibitemShut {NoStop}%
\bibitem [{\citenamefont {Lane}\ and\ \citenamefont {Ersson}(2016)}]{lane2016culture}%
  \BibitemOpen
  \bibfield  {author} {\bibinfo {author} {\bibfnamefont {J.-E.}\ \bibnamefont {Lane}}\ and\ \bibinfo {author} {\bibfnamefont {S.}~\bibnamefont {Ersson}},\ }\href {https://doi.org/10.4324/9781315575452} {\emph {\bibinfo {title} {Culture and {{Politics}}: {{A Comparative Approach}}}}},\ \bibinfo {edition} {2nd}\ ed.\ (\bibinfo  {publisher} {{Routledge}},\ \bibinfo {address} {{London}},\ \bibinfo {year} {2016})\BibitemShut {NoStop}%
\bibitem [{\citenamefont {Odum}(1936)}]{OdumSouthernRegions1936}%
  \BibitemOpen
  \bibfield  {author} {\bibinfo {author} {\bibfnamefont {H.~W.}\ \bibnamefont {Odum}},\ }\href@noop {} {\emph {\bibinfo {title} {Southern Regions of the United States}}}\ (\bibinfo  {publisher} {{Univ. North Carolina Press}},\ \bibinfo {address} {{Chapel Hill, NC}},\ \bibinfo {year} {1936})\BibitemShut {NoStop}%
\bibitem [{\citenamefont {Elazar}(1970)}]{ElazarCitiesPrairie1970}%
  \BibitemOpen
  \bibfield  {author} {\bibinfo {author} {\bibfnamefont {D.~J.}\ \bibnamefont {Elazar}},\ }\href@noop {} {\emph {\bibinfo {title} {Cities of the Prairie: The Metropolitan Frontier and American Politics}}}\ (\bibinfo  {publisher} {{Basic Books}},\ \bibinfo {address} {{New York}},\ \bibinfo {year} {1970})\BibitemShut {NoStop}%
\bibitem [{\citenamefont {Zelinsky}(1992)}]{ZelinskyCulturalGeography1992}%
  \BibitemOpen
  \bibfield  {author} {\bibinfo {author} {\bibfnamefont {W.}~\bibnamefont {Zelinsky}},\ }\href@noop {} {\emph {\bibinfo {title} {The Cultural Geography of the United States}}}\ (\bibinfo  {publisher} {{Prentice Hall}},\ \bibinfo {address} {{Englewood Cliffs}},\ \bibinfo {year} {1992})\BibitemShut {NoStop}%
\bibitem [{\citenamefont {Gastil}(1975)}]{GastilCulturalRegions1975}%
  \BibitemOpen
  \bibfield  {author} {\bibinfo {author} {\bibfnamefont {R.~D.}\ \bibnamefont {Gastil}},\ }\href@noop {} {\emph {\bibinfo {title} {Cultural Regions of the United States}}}\ (\bibinfo  {publisher} {{University of Washington Press}},\ \bibinfo {address} {{Seattle}},\ \bibinfo {year} {1975})\BibitemShut {NoStop}%
\bibitem [{\citenamefont {Garreau}(1996)}]{GarreauNineNations1996}%
  \BibitemOpen
  \bibfield  {author} {\bibinfo {author} {\bibfnamefont {J.}~\bibnamefont {Garreau}},\ }\href@noop {} {\emph {\bibinfo {title} {The Nine Nations of North America}}}\ (\bibinfo  {publisher} {{Houghton Mifflin Company}},\ \bibinfo {address} {{Boston}},\ \bibinfo {year} {1996})\BibitemShut {NoStop}%
\bibitem [{\citenamefont {Lieske}(1993)}]{LieskeRegionalSubcultures1993}%
  \BibitemOpen
  \bibfield  {author} {\bibinfo {author} {\bibfnamefont {J.}~\bibnamefont {Lieske}},\ }\bibfield  {title} {\bibinfo {title} {Regional subcultures of the united states},\ }\href {https://doi.org/10.2307/2131941} {\bibfield  {journal} {\bibinfo  {journal} {The Journal of Politics}\ }\textbf {\bibinfo {volume} {55}},\ \bibinfo {pages} {888} (\bibinfo {year} {1993})}\BibitemShut {NoStop}%
\bibitem [{\citenamefont {Woodard}(2012)}]{WoodardAmericanNations2012}%
  \BibitemOpen
  \bibfield  {author} {\bibinfo {author} {\bibfnamefont {C.}~\bibnamefont {Woodard}},\ }\href {https://www.worldcat.org/title/american-nations-a-history-of-the-eleven-rival-regional-cultures-of-north-america/oclc/810122408&referer=brief_results} {\emph {\bibinfo {title} {American Nations: A History of the Eleven Rival Regional Cultures of North America}}}\ (\bibinfo  {publisher} {{Penguin Books}},\ \bibinfo {address} {{New York NY}},\ \bibinfo {year} {2012})\BibitemShut {NoStop}%
\bibitem [{\citenamefont {Fischer}(1989)}]{FischerAlbionSeed1989}%
  \BibitemOpen
  \bibfield  {author} {\bibinfo {author} {\bibfnamefont {D.~H.}\ \bibnamefont {Fischer}},\ }\href@noop {} {\emph {\bibinfo {title} {Albion's Seed}}}\ (\bibinfo  {publisher} {{Oxford University Press}},\ \bibinfo {address} {{Oxford, UK}},\ \bibinfo {year} {1989})\BibitemShut {NoStop}%
\bibitem [{\citenamefont {Kramsch}(2014)}]{kramsch2014language}%
  \BibitemOpen
  \bibfield  {author} {\bibinfo {author} {\bibfnamefont {C.}~\bibnamefont {Kramsch}},\ }\bibfield  {title} {\bibinfo {title} {Language and culture},\ }\href {https://doi.org/10.1075/aila.27.02kra} {\bibfield  {journal} {\bibinfo  {journal} {AILA review}\ }\textbf {\bibinfo {volume} {27}},\ \bibinfo {pages} {30} (\bibinfo {year} {2014})}\BibitemShut {NoStop}%
\bibitem [{\citenamefont {Nguyen}\ \emph {et~al.}(2016)\citenamefont {Nguyen}, \citenamefont {Do{\u g}ru{\"o}z}, \citenamefont {Ros{\'e}},\ and\ \citenamefont {{de Jong}}}]{NguyenComputationalSociolinguistics2016}%
  \BibitemOpen
  \bibfield  {author} {\bibinfo {author} {\bibfnamefont {D.}~\bibnamefont {Nguyen}}, \bibinfo {author} {\bibfnamefont {A.~S.}\ \bibnamefont {Do{\u g}ru{\"o}z}}, \bibinfo {author} {\bibfnamefont {C.~P.}\ \bibnamefont {Ros{\'e}}},\ and\ \bibinfo {author} {\bibfnamefont {F.}~\bibnamefont {{de Jong}}},\ }\bibfield  {title} {\bibinfo {title} {Computational sociolinguistics: A survey},\ }\href {https://doi.org/10.1162/COLI_a_00258} {\bibfield  {journal} {\bibinfo  {journal} {Computational Linguistics}\ }\textbf {\bibinfo {volume} {42}},\ \bibinfo {pages} {537} (\bibinfo {year} {2016})}\BibitemShut {NoStop}%
\bibitem [{\citenamefont {Grieve}\ \emph {et~al.}(2011)\citenamefont {Grieve}, \citenamefont {Speelman},\ and\ \citenamefont {Geeraerts}}]{GrieveStatisticalMethod2011}%
  \BibitemOpen
  \bibfield  {author} {\bibinfo {author} {\bibfnamefont {J.}~\bibnamefont {Grieve}}, \bibinfo {author} {\bibfnamefont {D.}~\bibnamefont {Speelman}},\ and\ \bibinfo {author} {\bibfnamefont {D.}~\bibnamefont {Geeraerts}},\ }\bibfield  {title} {\bibinfo {title} {A statistical method for the identification and aggregation of regional linguistic variation},\ }\href {https://doi.org/10.1017/S095439451100007X} {\bibfield  {journal} {\bibinfo  {journal} {Language Variation and Change}\ }\textbf {\bibinfo {volume} {23}},\ \bibinfo {pages} {193} (\bibinfo {year} {2011})}\BibitemShut {NoStop}%
\bibitem [{\citenamefont {Eisenstein}\ \emph {et~al.}(2014)\citenamefont {Eisenstein}, \citenamefont {O'Connor}, \citenamefont {Smith},\ and\ \citenamefont {Xing}}]{EisensteinDiffusionLexical2014}%
  \BibitemOpen
  \bibfield  {author} {\bibinfo {author} {\bibfnamefont {J.}~\bibnamefont {Eisenstein}}, \bibinfo {author} {\bibfnamefont {B.}~\bibnamefont {O'Connor}}, \bibinfo {author} {\bibfnamefont {N.~A.}\ \bibnamefont {Smith}},\ and\ \bibinfo {author} {\bibfnamefont {E.~P.}\ \bibnamefont {Xing}},\ }\bibfield  {title} {\bibinfo {title} {Diffusion of lexical change in social media},\ }\href {https://doi.org/10.1371/journal.pone.0113114} {\bibfield  {journal} {\bibinfo  {journal} {PLoS ONE}\ }\textbf {\bibinfo {volume} {9}},\ \bibinfo {pages} {e113114} (\bibinfo {year} {2014})}\BibitemShut {NoStop}%
\bibitem [{\citenamefont {Gon{\c c}alves}\ and\ \citenamefont {Sanchez}(2014)}]{GoncalvesCrowdsourcingDialect2014}%
  \BibitemOpen
  \bibfield  {author} {\bibinfo {author} {\bibfnamefont {B.}~\bibnamefont {Gon{\c c}alves}}\ and\ \bibinfo {author} {\bibfnamefont {D.}~\bibnamefont {Sanchez}},\ }\bibfield  {title} {\bibinfo {title} {Crowdsourcing dialect characterization through {{Twitter}}},\ }\href {https://doi.org/10.1371/journal.pone.0112074} {\bibfield  {journal} {\bibinfo  {journal} {PLOS ONE}\ }\textbf {\bibinfo {volume} {9}},\ \bibinfo {pages} {e112074} (\bibinfo {year} {2014})}\BibitemShut {NoStop}%
\bibitem [{\citenamefont {Huang}\ \emph {et~al.}(2016)\citenamefont {Huang}, \citenamefont {Guo}, \citenamefont {Kasakoff},\ and\ \citenamefont {Grieve}}]{HuangUnderstandingRegional2016}%
  \BibitemOpen
  \bibfield  {author} {\bibinfo {author} {\bibfnamefont {Y.}~\bibnamefont {Huang}}, \bibinfo {author} {\bibfnamefont {D.}~\bibnamefont {Guo}}, \bibinfo {author} {\bibfnamefont {A.}~\bibnamefont {Kasakoff}},\ and\ \bibinfo {author} {\bibfnamefont {J.}~\bibnamefont {Grieve}},\ }\bibfield  {title} {\bibinfo {title} {Understanding {{U}}.{{S}}. regional linguistic variation with twitter data analysis},\ }\href {https://doi.org/10.1016/j.compenvurbsys.2015.12.003} {\bibfield  {journal} {\bibinfo  {journal} {Computers, Environment and Urban Systems}\ }\textbf {\bibinfo {volume} {59}},\ \bibinfo {pages} {244} (\bibinfo {year} {2016})}\BibitemShut {NoStop}%
\bibitem [{\citenamefont {Grieve}(2016)}]{GrieveRegionalVariation2016}%
  \BibitemOpen
  \bibfield  {author} {\bibinfo {author} {\bibfnamefont {J.}~\bibnamefont {Grieve}},\ }\href {https://doi.org/10.1017/CBO9781139506137} {\emph {\bibinfo {title} {Regional Variation in Written American English}}}\ (\bibinfo  {publisher} {{Cambridge University Press}},\ \bibinfo {year} {2016})\BibitemShut {NoStop}%
\bibitem [{\citenamefont {Donoso}\ and\ \citenamefont {S{\'a}nchez}(2017)}]{Donoso2017}%
  \BibitemOpen
  \bibfield  {author} {\bibinfo {author} {\bibfnamefont {G.}~\bibnamefont {Donoso}}\ and\ \bibinfo {author} {\bibfnamefont {D.}~\bibnamefont {S{\'a}nchez}},\ }\bibfield  {title} {\bibinfo {title} {Dialectometric analysis of language variation in {{Twitter}}},\ }in\ \href {https://doi.org/10.18653/v1/W17-1202} {\emph {\bibinfo {booktitle} {Proceedings of the Fourth Workshop on {{NLP}} for Similar Languages, Varieties and Dialects ({{VarDial}})}}}\ (\bibinfo {year} {2017})\ pp.\ \bibinfo {pages} {16--25}\BibitemShut {NoStop}%
\bibitem [{\citenamefont {Gon{\c c}alves}\ \emph {et~al.}(2018)\citenamefont {Gon{\c c}alves}, \citenamefont {{Loureiro-Porto}}, \citenamefont {Ramasco},\ and\ \citenamefont {S{\'a}nchez}}]{GoncalvesMappingAmericanization2018}%
  \BibitemOpen
  \bibfield  {author} {\bibinfo {author} {\bibfnamefont {B.}~\bibnamefont {Gon{\c c}alves}}, \bibinfo {author} {\bibfnamefont {L.}~\bibnamefont {{Loureiro-Porto}}}, \bibinfo {author} {\bibfnamefont {J.~J.}\ \bibnamefont {Ramasco}},\ and\ \bibinfo {author} {\bibfnamefont {D.}~\bibnamefont {S{\'a}nchez}},\ }\bibfield  {title} {\bibinfo {title} {Mapping the americanization of {{English}} in space and time},\ }\href {https://doi.org/10.1371/journal.pone.0197741} {\bibfield  {journal} {\bibinfo  {journal} {PLOS ONE}\ }\textbf {\bibinfo {volume} {13}},\ \bibinfo {pages} {e0197741} (\bibinfo {year} {2018})}\BibitemShut {NoStop}%
\bibitem [{\citenamefont {Abitbol}\ \emph {et~al.}(2018)\citenamefont {Abitbol}, \citenamefont {Karsai}, \citenamefont {Magu{\'e}}, \citenamefont {Chevrot},\ and\ \citenamefont {Fleury}}]{AbitbolSocioeconomicDependencies2018}%
  \BibitemOpen
  \bibfield  {author} {\bibinfo {author} {\bibfnamefont {J.~L.}\ \bibnamefont {Abitbol}}, \bibinfo {author} {\bibfnamefont {M.}~\bibnamefont {Karsai}}, \bibinfo {author} {\bibfnamefont {J.~P.}\ \bibnamefont {Magu{\'e}}}, \bibinfo {author} {\bibfnamefont {J.~P.}\ \bibnamefont {Chevrot}},\ and\ \bibinfo {author} {\bibfnamefont {E.}~\bibnamefont {Fleury}},\ }\bibfield  {title} {\bibinfo {title} {Socioeconomic dependencies of linguistic patterns in {{Twitter}}: A multivariate analysis},\ }in\ \href {https://doi.org/10.1145/3178876.3186011} {\emph {\bibinfo {booktitle} {The {{Web Conference}} 2018 - {{Proceedings}} of the {{World Wide Web Conference}}, {{WWW}} 2018}}}\ (\bibinfo {year} {2018})\ pp.\ \bibinfo {pages} {1125--1134},\ \Eprint {https://arxiv.org/abs/1804.01155} {arXiv:1804.01155} \BibitemShut {NoStop}%
\bibitem [{\citenamefont {Grieve}\ \emph {et~al.}(2019)\citenamefont {Grieve}, \citenamefont {Montgomery}, \citenamefont {Nini}, \citenamefont {Murakami},\ and\ \citenamefont {Guo}}]{GrieveMappingLexical2019}%
  \BibitemOpen
  \bibfield  {author} {\bibinfo {author} {\bibfnamefont {J.}~\bibnamefont {Grieve}}, \bibinfo {author} {\bibfnamefont {C.}~\bibnamefont {Montgomery}}, \bibinfo {author} {\bibfnamefont {A.}~\bibnamefont {Nini}}, \bibinfo {author} {\bibfnamefont {A.}~\bibnamefont {Murakami}},\ and\ \bibinfo {author} {\bibfnamefont {D.}~\bibnamefont {Guo}},\ }\bibfield  {title} {\bibinfo {title} {Mapping lexical dialect variation in british {{English}} using {{Twitter}}},\ }\href {https://doi.org/10.3389/frai.2019.00011} {\bibfield  {journal} {\bibinfo  {journal} {Frontiers in Artificial Intelligence}\ }\textbf {\bibinfo {volume} {2}},\ \bibinfo {pages} {11} (\bibinfo {year} {2019})}\BibitemShut {NoStop}%
\bibitem [{\citenamefont {Koylu}(2018)}]{KoyluUncoveringGeoSocial2018}%
  \BibitemOpen
  \bibfield  {author} {\bibinfo {author} {\bibfnamefont {C.}~\bibnamefont {Koylu}},\ }\bibfield  {title} {\bibinfo {title} {Uncovering {{Geo-Social Semantics}} from the {{Twitter Mention Network}}: {{An Integrated Approach Using Spatial Network Smoothing}} and {{Topic Modeling}}},\ }in\ \href {https://doi.org/10.1007/978-3-319-73247-3_9} {\emph {\bibinfo {booktitle} {Human {{Dynamics Research}} in {{Smart}} and {{Connected Communities}}}}},\ \bibinfo {series and number} {Human {{Dynamics}} in {{Smart Cities}}},\ \bibinfo {editor} {edited by\ \bibinfo {editor} {\bibfnamefont {S.-L.}\ \bibnamefont {Shaw}}\ and\ \bibinfo {editor} {\bibfnamefont {D.}~\bibnamefont {Sui}}}\ (\bibinfo  {publisher} {{Springer International Publishing}},\ \bibinfo {address} {{Cham}},\ \bibinfo {year} {2018})\ pp.\ \bibinfo {pages} {163--179}\BibitemShut {NoStop}%
\bibitem [{\citenamefont {Funkner}\ \emph {et~al.}(2021)\citenamefont {Funkner}, \citenamefont {Elkhovskaya}, \citenamefont {Lenivtceva}, \citenamefont {Egorov}, \citenamefont {Kshenin},\ and\ \citenamefont {Khrulkov}}]{FunknerGeographicalTopic2021}%
  \BibitemOpen
  \bibfield  {author} {\bibinfo {author} {\bibfnamefont {A.~A.}\ \bibnamefont {Funkner}}, \bibinfo {author} {\bibfnamefont {L.~O.}\ \bibnamefont {Elkhovskaya}}, \bibinfo {author} {\bibfnamefont {I.~D.}\ \bibnamefont {Lenivtceva}}, \bibinfo {author} {\bibfnamefont {M.~P.}\ \bibnamefont {Egorov}}, \bibinfo {author} {\bibfnamefont {A.~D.}\ \bibnamefont {Kshenin}},\ and\ \bibinfo {author} {\bibfnamefont {A.~A.}\ \bibnamefont {Khrulkov}},\ }\bibfield  {title} {\bibinfo {title} {Geographical {{Topic Modelling}} on {{Spatial Social Network Data}}},\ }\href {https://doi.org/10.1016/j.procs.2021.10.003} {\bibfield  {journal} {\bibinfo  {journal} {Procedia Computer Science}\ }\bibinfo {series} {10th {{International Young Scientists Conference}} in {{Computational Science}}, {{YSC2021}}, 28 {{June}} \textendash{} 2 {{July}}, 2021},\ \textbf {\bibinfo {volume} {193}},\ \bibinfo {pages} {22} (\bibinfo {year} {2021})}\BibitemShut {NoStop}%
\bibitem [{\citenamefont {Mislove}\ \emph {et~al.}(2011)\citenamefont {Mislove}, \citenamefont {Lehmann}, \citenamefont {Ahn}, \citenamefont {Onnela},\ and\ \citenamefont {Rosenquist}}]{MisloveUnderstandingDemographics2011}%
  \BibitemOpen
  \bibfield  {author} {\bibinfo {author} {\bibfnamefont {A.}~\bibnamefont {Mislove}}, \bibinfo {author} {\bibfnamefont {S.}~\bibnamefont {Lehmann}}, \bibinfo {author} {\bibfnamefont {Y.-Y.}\ \bibnamefont {Ahn}}, \bibinfo {author} {\bibfnamefont {J.-P.}\ \bibnamefont {Onnela}},\ and\ \bibinfo {author} {\bibfnamefont {J.~N.}\ \bibnamefont {Rosenquist}},\ }\bibfield  {title} {\bibinfo {title} {Understanding the demographics of {{Twitter}} users},\ }in\ \href {https://orbit.dtu.dk/en/publications/understanding-the-demographics-of-twitter-users} {\emph {\bibinfo {booktitle} {Proceedings of the {{International AAAI Conference}} on {{Web}} and {{Social Media}}}}},\ Vol.~\bibinfo {volume} {5}\ (\bibinfo  {publisher} {{AAAI Press}},\ \bibinfo {address} {{Barcelona}},\ \bibinfo {year} {2011})\ pp.\ \bibinfo {pages} {554--557}\BibitemShut {NoStop}%
\bibitem [{\citenamefont {Pavalanathan}\ and\ \citenamefont {Eisenstein}(2015)}]{PavalanathanConfoundsConsequences2015}%
  \BibitemOpen
  \bibfield  {author} {\bibinfo {author} {\bibfnamefont {U.}~\bibnamefont {Pavalanathan}}\ and\ \bibinfo {author} {\bibfnamefont {J.}~\bibnamefont {Eisenstein}},\ }\bibfield  {title} {\bibinfo {title} {Confounds and consequences in geotagged {{Twitter}} data},\ }in\ \href {https://doi.org/10.18653/v1/d15-1256} {\emph {\bibinfo {booktitle} {Proceedings of the 2015 {{Conference}} on {{Empirical Methods}} in {{Natural Language Processing}}}}}\ (\bibinfo  {publisher} {{Association for Computational Linguistics (ACL)}},\ \bibinfo {address} {{Lisbon}},\ \bibinfo {year} {2015})\ pp.\ \bibinfo {pages} {2138--2148},\ \Eprint {https://arxiv.org/abs/1506.02275} {arXiv:1506.02275} \BibitemShut {NoStop}%
\bibitem [{\citenamefont {Steiger}\ \emph {et~al.}(2015)\citenamefont {Steiger}, \citenamefont {De~Albuquerque},\ and\ \citenamefont {Zipf}}]{steiger2015advanced}%
  \BibitemOpen
  \bibfield  {author} {\bibinfo {author} {\bibfnamefont {E.}~\bibnamefont {Steiger}}, \bibinfo {author} {\bibfnamefont {J.~P.}\ \bibnamefont {De~Albuquerque}},\ and\ \bibinfo {author} {\bibfnamefont {A.}~\bibnamefont {Zipf}},\ }\bibfield  {title} {\bibinfo {title} {An advanced systematic literature review on spatiotemporal analyses of {{Twitter}} data},\ }\href {https://doi.org/10.1111/tgis.12132} {\bibfield  {journal} {\bibinfo  {journal} {Transactions in GIS}\ }\textbf {\bibinfo {volume} {19}},\ \bibinfo {pages} {809} (\bibinfo {year} {2015})}\BibitemShut {NoStop}%
\bibitem [{\citenamefont {Diaz}\ \emph {et~al.}(2016)\citenamefont {Diaz}, \citenamefont {Gamon}, \citenamefont {Hofman}, \citenamefont {K{\i}c{\i}man},\ and\ \citenamefont {Rothschild}}]{diaz2016online}%
  \BibitemOpen
  \bibfield  {author} {\bibinfo {author} {\bibfnamefont {F.}~\bibnamefont {Diaz}}, \bibinfo {author} {\bibfnamefont {M.}~\bibnamefont {Gamon}}, \bibinfo {author} {\bibfnamefont {J.~M.}\ \bibnamefont {Hofman}}, \bibinfo {author} {\bibfnamefont {E.}~\bibnamefont {K{\i}c{\i}man}},\ and\ \bibinfo {author} {\bibfnamefont {D.}~\bibnamefont {Rothschild}},\ }\bibfield  {title} {\bibinfo {title} {Online and social media data as an imperfect continuous panel survey},\ }\href {https://doi.org/10.1371/journal.pone.0145406} {\bibfield  {journal} {\bibinfo  {journal} {PLOS ONE}\ }\textbf {\bibinfo {volume} {11}},\ \bibinfo {pages} {e0145406} (\bibinfo {year} {2016})}\BibitemShut {NoStop}%
\bibitem [{\citenamefont {Auxier}\ and\ \citenamefont {Anderson}(2021)}]{AuxierSocialMedia2021}%
  \BibitemOpen
  \bibfield  {author} {\bibinfo {author} {\bibfnamefont {B.}~\bibnamefont {Auxier}}\ and\ \bibinfo {author} {\bibfnamefont {M.}~\bibnamefont {Anderson}},\ }\href {https://www.pewresearch.org/internet/2021/04/07/social-media-use-in-2021/} {\emph {\bibinfo {title} {Social Media Use in 2021}}},\ \bibinfo {type} {Tech. Rep.}\ (\bibinfo  {institution} {{Pew Research Center}},\ \bibinfo {year} {2021})\BibitemShut {NoStop}%
\bibitem [{\citenamefont {{Al-Rfou}}\ and\ \citenamefont {Solomon}(2014)}]{Al-RfouPythonBindings2014}%
  \BibitemOpen
  \bibfield  {author} {\bibinfo {author} {\bibfnamefont {R.}~\bibnamefont {{Al-Rfou}}}\ and\ \bibinfo {author} {\bibfnamefont {B.}~\bibnamefont {Solomon}},\ }\href {https://github.com/aboSamoor/pycld2} {\bibinfo {title} {Python bindings for the compact language detector 2}} (\bibinfo {year} {2014})\BibitemShut {NoStop}%
\bibitem [{\citenamefont {Ord}\ and\ \citenamefont {Getis}(1995)}]{OrdLocalSpatial1995}%
  \BibitemOpen
  \bibfield  {author} {\bibinfo {author} {\bibfnamefont {J.~K.}\ \bibnamefont {Ord}}\ and\ \bibinfo {author} {\bibfnamefont {A.}~\bibnamefont {Getis}},\ }\bibfield  {title} {\bibinfo {title} {Local spatial autocorrelation statistics: Distributional issues and an application},\ }\href {https://doi.org/10.1111/J.1538-4632.1995.TB00912.X} {\bibfield  {journal} {\bibinfo  {journal} {Geographical Analysis}\ }\textbf {\bibinfo {volume} {27}},\ \bibinfo {pages} {286} (\bibinfo {year} {1995})}\BibitemShut {NoStop}%
\bibitem [{\citenamefont {Wold}\ \emph {et~al.}(1987)\citenamefont {Wold}, \citenamefont {Esbensen},\ and\ \citenamefont {Geladi}}]{WoldPrincipalComponent1987}%
  \BibitemOpen
  \bibfield  {author} {\bibinfo {author} {\bibfnamefont {S.}~\bibnamefont {Wold}}, \bibinfo {author} {\bibfnamefont {K.}~\bibnamefont {Esbensen}},\ and\ \bibinfo {author} {\bibfnamefont {P.}~\bibnamefont {Geladi}},\ }\bibfield  {title} {\bibinfo {title} {Principal component analysis},\ }\href {https://doi.org/10.1016/0169-7439(87)80084-9} {\bibfield  {journal} {\bibinfo  {journal} {Chemometrics and Intelligent Laboratory Systems}\ }\textbf {\bibinfo {volume} {2}},\ \bibinfo {pages} {37} (\bibinfo {year} {1987})}\BibitemShut {NoStop}%
\bibitem [{\citenamefont {Arun}\ \emph {et~al.}(2010)\citenamefont {Arun}, \citenamefont {Suresh}, \citenamefont {Veni~Madhavan},\ and\ \citenamefont {Narasimha~Murthy}}]{ArunFindingNatural2010}%
  \BibitemOpen
  \bibfield  {author} {\bibinfo {author} {\bibfnamefont {R.}~\bibnamefont {Arun}}, \bibinfo {author} {\bibfnamefont {V.}~\bibnamefont {Suresh}}, \bibinfo {author} {\bibfnamefont {C.~E.}\ \bibnamefont {Veni~Madhavan}},\ and\ \bibinfo {author} {\bibfnamefont {M.~N.}\ \bibnamefont {Narasimha~Murthy}},\ }\bibfield  {title} {\bibinfo {title} {On finding the natural number of topics with latent dirichlet allocation: Some observations},\ }in\ \href {https://doi.org/10.1007/978-3-642-13657-3_43} {\emph {\bibinfo {booktitle} {Proceedings of the 14th {{Pacific-Asia}} Conference on {{Advances}} in {{Knowledge Discovery}} and {{Data Mining}} - {{Volume Part I}}}}},\ \bibinfo {series and number} {{{PAKDD}}'10}\ (\bibinfo  {publisher} {{Springer-Verlag}},\ \bibinfo {address} {{Berlin, Heidelberg}},\ \bibinfo {year} {2010})\ pp.\ \bibinfo {pages} {391--402}\BibitemShut {NoStop}%
\bibitem [{\citenamefont {Hasan}\ \emph {et~al.}(2021)\citenamefont {Hasan}, \citenamefont {Rahman}, \citenamefont {Karim}, \citenamefont {Khan},\ and\ \citenamefont {Islam}}]{HasanNormalizedApproach2021}%
  \BibitemOpen
  \bibfield  {author} {\bibinfo {author} {\bibfnamefont {M.}~\bibnamefont {Hasan}}, \bibinfo {author} {\bibfnamefont {A.}~\bibnamefont {Rahman}}, \bibinfo {author} {\bibfnamefont {M.~R.}\ \bibnamefont {Karim}}, \bibinfo {author} {\bibfnamefont {M.~S.~I.}\ \bibnamefont {Khan}},\ and\ \bibinfo {author} {\bibfnamefont {M.~J.}\ \bibnamefont {Islam}},\ }\bibfield  {title} {\bibinfo {title} {Normalized {{Approach}} to {{Find Optimal Number}} of {{Topics}} in {{Latent Dirichlet Allocation}} ({{LDA}})},\ }in\ \href {https://doi.org/10.1007/978-981-33-4673-4_27} {\emph {\bibinfo {booktitle} {Proceedings of {{International Conference}} on {{Trends}} in {{Computational}} and {{Cognitive Engineering}}}}},\ \bibinfo {series and number} {Advances in {{Intelligent Systems}} and {{Computing}}},\ \bibinfo {editor} {edited by\ \bibinfo {editor} {\bibfnamefont {M.~S.}\ \bibnamefont {Kaiser}}, \bibinfo {editor} {\bibfnamefont {A.}~\bibnamefont {Bandyopadhyay}}, \bibinfo {editor} {\bibfnamefont {M.}~\bibnamefont {Mahmud}},\ and\ \bibinfo {editor} {\bibfnamefont {K.}~\bibnamefont {Ray}}}\ (\bibinfo  {publisher} {{Springer}},\ \bibinfo {address} {{Singapore}},\ \bibinfo {year} {2021})\ pp.\ \bibinfo {pages} {341--354}\BibitemShut {NoStop}%
\bibitem [{\citenamefont {Frontier}(1976)}]{FrontierEtudeDecroissance1976}%
  \BibitemOpen
  \bibfield  {author} {\bibinfo {author} {\bibfnamefont {S.}~\bibnamefont {Frontier}},\ }\bibfield  {title} {\bibinfo {title} {\'etude de la d\'ecroissance des valeurs propres dans une analyse en composantes principales: Comparaison avec le mod\`ele du b\^aton bris\'e},\ }\href {https://doi.org/10.1016/0022-0981(76)90076-9} {\bibfield  {journal} {\bibinfo  {journal} {Journal of Experimental Marine Biology and Ecology}\ }\textbf {\bibinfo {volume} {25}},\ \bibinfo {pages} {67} (\bibinfo {year} {1976})}\BibitemShut {NoStop}%
\bibitem [{\citenamefont {Jackson}(1993)}]{JacksonStoppingRules1993}%
  \BibitemOpen
  \bibfield  {author} {\bibinfo {author} {\bibfnamefont {D.~A.}\ \bibnamefont {Jackson}},\ }\bibfield  {title} {\bibinfo {title} {Stopping rules in principal components analysis: A comparison of heuristical and statistical approaches},\ }\href {https://doi.org/10.2307/1939574} {\bibfield  {journal} {\bibinfo  {journal} {Ecology}\ }\textbf {\bibinfo {volume} {74}},\ \bibinfo {pages} {2204} (\bibinfo {year} {1993})}\BibitemShut {NoStop}%
\bibitem [{\citenamefont {Everitt}\ \emph {et~al.}(2011)\citenamefont {Everitt}, \citenamefont {Landau}, \citenamefont {Leese},\ and\ \citenamefont {Stahl}}]{EverittClusterAnalysis2011}%
  \BibitemOpen
  \bibfield  {author} {\bibinfo {author} {\bibfnamefont {B.~S.}\ \bibnamefont {Everitt}}, \bibinfo {author} {\bibfnamefont {S.}~\bibnamefont {Landau}}, \bibinfo {author} {\bibfnamefont {M.}~\bibnamefont {Leese}},\ and\ \bibinfo {author} {\bibfnamefont {D.}~\bibnamefont {Stahl}},\ }\href {https://www.wiley.com/en-us/Cluster+Analysis%2C+5th+Edition-p-9780470749913} {\emph {\bibinfo {title} {Cluster Analysis}}}\ (\bibinfo  {publisher} {{John Wiley \& Sons}},\ \bibinfo {address} {{Wiley, Chichester, UK}},\ \bibinfo {year} {2011})\BibitemShut {NoStop}%
\bibitem [{\citenamefont {Rousseeuw}(1987)}]{RousseeuwSilhouettesGraphical1987}%
  \BibitemOpen
  \bibfield  {author} {\bibinfo {author} {\bibfnamefont {P.~J.}\ \bibnamefont {Rousseeuw}},\ }\bibfield  {title} {\bibinfo {title} {Silhouettes: A graphical aid to the interpretation and validation of cluster analysis},\ }\href {https://doi.org/10.1016/0377-0427(87)90125-7} {\bibfield  {journal} {\bibinfo  {journal} {Journal of Computational and Applied Mathematics}\ }\textbf {\bibinfo {volume} {20}},\ \bibinfo {pages} {53} (\bibinfo {year} {1987})}\BibitemShut {NoStop}%
\bibitem [{\citenamefont {Vanderbeck}\ and\ \citenamefont {Dunkley}(2003)}]{vanderbeck2003young}%
  \BibitemOpen
  \bibfield  {author} {\bibinfo {author} {\bibfnamefont {R.~M.}\ \bibnamefont {Vanderbeck}}\ and\ \bibinfo {author} {\bibfnamefont {C.~M.}\ \bibnamefont {Dunkley}},\ }\bibfield  {title} {\bibinfo {title} {Young people's narratives of rural-urban difference},\ }\href {https://doi.org/10.1080/14733280302192} {\bibfield  {journal} {\bibinfo  {journal} {Children's geographies}\ }\textbf {\bibinfo {volume} {1}},\ \bibinfo {pages} {241} (\bibinfo {year} {2003})}\BibitemShut {NoStop}%
\bibitem [{\citenamefont {Gelman}(2009)}]{GelmanRedState2009}%
  \BibitemOpen
  \bibfield  {author} {\bibinfo {author} {\bibfnamefont {A.}~\bibnamefont {Gelman}},\ }\href {https://doi.org/10.1515/9781400832118} {\emph {\bibinfo {title} {Red State, Blue State, Rich State, Poor State: Why Americans Vote the Way They Do}}}\ (\bibinfo  {publisher} {{Princeton University Press}},\ \bibinfo {address} {{Princeton}},\ \bibinfo {year} {2009})\BibitemShut {NoStop}%
\bibitem [{\citenamefont {Bochkarev}\ \emph {et~al.}(2015)\citenamefont {Bochkarev}, \citenamefont {Shevlyakova},\ and\ \citenamefont {Solovyev}}]{BochkarevAverageWord2015}%
  \BibitemOpen
  \bibfield  {author} {\bibinfo {author} {\bibfnamefont {V.~V.}\ \bibnamefont {Bochkarev}}, \bibinfo {author} {\bibfnamefont {A.~V.}\ \bibnamefont {Shevlyakova}},\ and\ \bibinfo {author} {\bibfnamefont {V.~D.}\ \bibnamefont {Solovyev}},\ }\bibfield  {title} {\bibinfo {title} {The {{Average Word Length Dynamics}} as an {{Indicator}} of {{Cultural Changes}} in {{Society}}},\ }\href@noop {} {\bibfield  {journal} {\bibinfo  {journal} {Social Evolution \& History}\ }\textbf {\bibinfo {volume} {14}},\ \bibinfo {pages} {153} (\bibinfo {year} {2015})}\BibitemShut {NoStop}%
\bibitem [{\citenamefont {Bentley}\ \emph {et~al.}(2014)\citenamefont {Bentley}, \citenamefont {Acerbi}, \citenamefont {Ormerod},\ and\ \citenamefont {Lampos}}]{BentleyBooksAverage2014}%
  \BibitemOpen
  \bibfield  {author} {\bibinfo {author} {\bibfnamefont {R.~A.}\ \bibnamefont {Bentley}}, \bibinfo {author} {\bibfnamefont {A.}~\bibnamefont {Acerbi}}, \bibinfo {author} {\bibfnamefont {P.}~\bibnamefont {Ormerod}},\ and\ \bibinfo {author} {\bibfnamefont {V.}~\bibnamefont {Lampos}},\ }\bibfield  {title} {\bibinfo {title} {Books {{Average Previous Decade}} of {{Economic Misery}}},\ }\href {https://doi.org/10.1371/journal.pone.0083147} {\bibfield  {journal} {\bibinfo  {journal} {PLOS ONE}\ }\textbf {\bibinfo {volume} {9}},\ \bibinfo {pages} {e83147} (\bibinfo {year} {2014})}\BibitemShut {NoStop}%
\bibitem [{\citenamefont {Karjus}\ \emph {et~al.}(2020)\citenamefont {Karjus}, \citenamefont {Blythe}, \citenamefont {Kirby},\ and\ \citenamefont {Smith}}]{KarjusQuantifyingDynamics2020}%
  \BibitemOpen
  \bibfield  {author} {\bibinfo {author} {\bibfnamefont {A.}~\bibnamefont {Karjus}}, \bibinfo {author} {\bibfnamefont {R.~A.}\ \bibnamefont {Blythe}}, \bibinfo {author} {\bibfnamefont {S.}~\bibnamefont {Kirby}},\ and\ \bibinfo {author} {\bibfnamefont {K.}~\bibnamefont {Smith}},\ }\bibfield  {title} {\bibinfo {title} {Quantifying the dynamics of topical fluctuations in language},\ }\href {https://doi.org/10.1163/22105832-01001200} {\bibfield  {journal} {\bibinfo  {journal} {Language Dynamics and Change}\ }\textbf {\bibinfo {volume} {10}},\ \bibinfo {pages} {86} (\bibinfo {year} {2020})}\BibitemShut {NoStop}%
\bibitem [{\citenamefont {Momeni}\ \emph {et~al.}(2018)\citenamefont {Momeni}, \citenamefont {Karunasekera}, \citenamefont {Goyal},\ and\ \citenamefont {Lerman}}]{MomeniModelingEvolution2018}%
  \BibitemOpen
  \bibfield  {author} {\bibinfo {author} {\bibfnamefont {E.}~\bibnamefont {Momeni}}, \bibinfo {author} {\bibfnamefont {S.}~\bibnamefont {Karunasekera}}, \bibinfo {author} {\bibfnamefont {P.}~\bibnamefont {Goyal}},\ and\ \bibinfo {author} {\bibfnamefont {K.}~\bibnamefont {Lerman}},\ }\bibfield  {title} {\bibinfo {title} {Modeling {{Evolution}} of {{Topics}} in {{Large-Scale Temporal Text Corpora}}},\ }in\ \href {https://www.aaai.org/ocs/index.php/ICWSM/ICWSM18/paper/view/17856} {\emph {\bibinfo {booktitle} {Twelfth {{International AAAI Conference}} on {{Web}} and {{Social Media}}}}}\ (\bibinfo {year} {2018})\ pp.\ \bibinfo {pages} {656--659}\BibitemShut {NoStop}%
\bibitem [{\citenamefont {Alshaabi}\ \emph {et~al.}(2021)\citenamefont {Alshaabi}, \citenamefont {Adams}, \citenamefont {Arnold}, \citenamefont {Minot}, \citenamefont {Dewhurst}, \citenamefont {Reagan}, \citenamefont {Danforth},\ and\ \citenamefont {Dodds}}]{AlshaabiStorywranglerMassive2021}%
  \BibitemOpen
  \bibfield  {author} {\bibinfo {author} {\bibfnamefont {T.}~\bibnamefont {Alshaabi}}, \bibinfo {author} {\bibfnamefont {J.~L.}\ \bibnamefont {Adams}}, \bibinfo {author} {\bibfnamefont {M.~V.}\ \bibnamefont {Arnold}}, \bibinfo {author} {\bibfnamefont {J.~R.}\ \bibnamefont {Minot}}, \bibinfo {author} {\bibfnamefont {D.~R.}\ \bibnamefont {Dewhurst}}, \bibinfo {author} {\bibfnamefont {A.~J.}\ \bibnamefont {Reagan}}, \bibinfo {author} {\bibfnamefont {C.~M.}\ \bibnamefont {Danforth}},\ and\ \bibinfo {author} {\bibfnamefont {P.~S.}\ \bibnamefont {Dodds}},\ }\bibfield  {title} {\bibinfo {title} {Storywrangler: {{A}} massive exploratorium for sociolinguistic, cultural, socioeconomic, and political timelines using {{Twitter}}},\ }\href {https://doi.org/10.1126/sciadv.abe6534} {\bibfield  {journal} {\bibinfo  {journal} {Science Advances}\ }\textbf {\bibinfo {volume} {7}},\ \bibinfo {pages} {eabe6534} (\bibinfo {year} {2021})}\BibitemShut {NoStop}%
\bibitem [{\citenamefont {Louf}(2023{\natexlab{a}})}]{LoufThomasWordsuse2023}%
  \BibitemOpen
  \bibfield  {author} {\bibinfo {author} {\bibfnamefont {T.}~\bibnamefont {Louf}},\ }\href {https://doi.org/10.6084/m9.figshare.20627034.v1} {\bibinfo {title} {{{words-use}}}} (\bibinfo {year} {2023}{\natexlab{a}})\BibitemShut {NoStop}%
\bibitem [{\citenamefont {Louf}(2023{\natexlab{b}})}]{LoufThomasWordCounts2023}%
  \BibitemOpen
  \bibfield  {author} {\bibinfo {author} {\bibfnamefont {T.}~\bibnamefont {Louf}},\ }\bibfield  {title} {\bibinfo {title} {Word counts per {{US}} county in geo-tagged {{Tweets}} posted between 2015 and 2021},\ }\href {https://doi.org/10.6084/m9.figshare.20630919.v1} {10.6084/m9.figshare.20630919.v1} (\bibinfo {year} {2023}{\natexlab{b}})\BibitemShut {NoStop}%
\end{thebibliography}%


\begin{thebibliography}{0}%
\makeatletter
\providecommand \@ifxundefined [1]{%
 \@ifx{#1\undefined}
}%
\providecommand \@ifnum [1]{%
 \ifnum #1\expandafter \@firstoftwo
 \else \expandafter \@secondoftwo
 \fi
}%
\providecommand \@ifx [1]{%
 \ifx #1\expandafter \@firstoftwo
 \else \expandafter \@secondoftwo
 \fi
}%
\providecommand \natexlab [1]{#1}%
\providecommand \enquote  [1]{``#1''}%
\providecommand \bibnamefont  [1]{#1}%
\providecommand \bibfnamefont [1]{#1}%
\providecommand \citenamefont [1]{#1}%
\providecommand \href@noop [0]{\@secondoftwo}%
\providecommand \href [0]{\begingroup \@sanitize@url \@href}%
\providecommand \@href[1]{\@@startlink{#1}\@@href}%
\providecommand \@@href[1]{\endgroup#1\@@endlink}%
\providecommand \@sanitize@url [0]{\catcode `\\12\catcode `\$12\catcode `\&12\catcode `\#12\catcode `\^12\catcode `\_12\catcode `\%12\relax}%
\providecommand \@@startlink[1]{}%
\providecommand \@@endlink[0]{}%
\providecommand \url  [0]{\begingroup\@sanitize@url \@url }%
\providecommand \@url [1]{\endgroup\@href {#1}{\urlprefix }}%
\providecommand \urlprefix  [0]{URL }%
\providecommand \Eprint [0]{\href }%
\providecommand \doibase [0]{https://doi.org/}%
\providecommand \selectlanguage [0]{\@gobble}%
\providecommand \bibinfo  [0]{\@secondoftwo}%
\providecommand \bibfield  [0]{\@secondoftwo}%
\providecommand \translation [1]{[#1]}%
\providecommand \BibitemOpen [0]{}%
\providecommand \bibitemStop [0]{}%
\providecommand \bibitemNoStop [0]{.\EOS\space}%
\providecommand \EOS [0]{\spacefactor3000\relax}%
\providecommand \BibitemShut  [1]{\csname bibitem#1\endcsname}%
\let\auto@bib@innerbib\@empty
\end{thebibliography}%

\end{document}


\title{Supplementary Information: American cultural regions mapped through the linguistic analysis of social media}

\author{Thomas Louf}
\email{thomaslouf@ifisc.uib-csic.es}
\affiliation{Institute for Cross-Disciplinary Physics and Complex Systems IFISC (UIB-CSIC), Palma de Mallorca, Spain}

\author{Bruno Gon\c calves}
\affiliation{ISI Foundation, Turin, Italy}

\author{Jos\'e J. Ramasco}
\affiliation{Institute for Cross-Disciplinary Physics and Complex Systems IFISC (UIB-CSIC), Palma de Mallorca, Spain}

\author{David S\'{a}nchez}
\email{david.sanchez@uib.es}
\affiliation{Institute for Cross-Disciplinary Physics and Complex Systems IFISC (UIB-CSIC), Palma de Mallorca, Spain}

\author{Jack Grieve}
\affiliation{Department of English Language and Linguistics, University of Birmingham, Birmingham, UK}
\affiliation{The Alan Turing Institute, London, UK}

\date{\today}

\maketitle

\section{Cultural regions' topical patterns}
In Fig.~4 of the main text, we present the cultural regions that follow from our analysis described in the Methods section. We hereafter show the most characteristic words for each region according to the specificity metric introduced in Eq.~(2) of the main text (the lists are by no means exhaustive). These words are classified into common lexical fields for the different cultural regions.

\begin{itemize}
  \item[\color{tol1}$\blacksquare$] Cluster 1
  \begin{itemize}
    \item[] \textbf{African-American variant forms and acronyms:} bih, frfr, stg, ian, otp, kno, ik, dnd, mfs, yo, sew, asf, wassup.
    \item[] \textbf{Fashion, music and food:} braids, waffle, grits, weave, lil, cookout, dreads, rappers, gucci.
    \item[] \textbf{Family and friends:} bruh, niggas, mama, dawg, shawty, auntie, folks, bae.
    \item[] \textbf{Location related:} atl, lsu, georgia, ga, carolina, saints, sc, falcons.
  \end{itemize}

  \item[\color{tol4}$\blacksquare$] Cluster 2
  \begin{itemize}
    \item[] \textbf{Sports:} win, teams, tourney, bracket, scoring, halftime, basketball, fouls, wrestling, innings, baseball.
    \item[] \textbf{Spectatorship:} cheering, announcers, congratulations, popcorn, hyped, concert, seats, autograph.
    \item[] \textbf{Teenage/school life:} girls, locker, choir, bffs, classroom, students, grader, alumni.
    \item[] \textbf{Location-related:} midwest, ohio, illinois, missouri, chicago, stl, minnesota, indiana, wisconsin, kc.
  \end{itemize}
  
  \item[\color{tol3}$\blacksquare$] Cluster 3
  \begin{itemize}
    \item[] \textbf{Natural features:} mountains, tree, snow, land, view, trail.
    \item[] \textbf{Outdoors activities:} adventures, spotter, trained, climb, trip.
    \item[] \textbf{Thoughts and emotions:} thought, think, cry, feel, crying, laughing, worry.
    \item[] \textbf{Personal communication:} said, understand, say, told, saying, heard, listen, trust, telling. (Importantly, these are negative features, i.e., it is a relatively low frequency of these words that is specific to the cluster.)
    \item[] \textbf{Location-related:} co, colorado, montana, utah, idaho, byu, mt, ut, va, wy.
  \end{itemize}
  
  \item[\color{tol2}$\blacksquare$] Cluster 4
  \begin{itemize}
    \item[] \textbf{Urban features:} freeway, homeless, traffic, ave.
    \item[] \textbf{Nationalities and origins:} puerto, dutch, latino, dominican, asian, immigrants, europe, rico, foreign, countries.
    \item[] \textbf{Violence:} violence, dangerous, attack, crime, insult, abusive, racist, threat.
    \item[] \textbf{Politics:} political, ban, nazi, indicted, protests, crisis, policies, presidency, elections, taxes, majority, obstruction.
    \item[] \textbf{Location related:} california, giants, nyc, ca, socal, cali, seattle, oregon, nj, sacramento, seahawks, sf, massachusetts, mets.
  \end{itemize}
  
  \item[\color{tol5}$\blacksquare$] Cluster 5
  \begin{itemize}
    \item[] \textbf{Mexican culture:} queso, mexican, tacos.
    \item[] \textbf{Variety features:} yalls, restroom, fixing, freaking, badass, highkey, fav, tripping.
    \item[] \textbf{Acronyms:} loml, sm, tbh, fml, idk, fm, fwm.
    \item[] \textbf{Emotions and relationships:} cry, heart, hurts, regret, friends, love, marry, boyfriend, laugh.
    \item[] \textbf{Location related:} whataburger, texas, tx, texans, dallas, cowboys, dfw, antonio, sa, spurs, mavs, as, ok, ou, okc, astros, rangers, austin.
  \end{itemize}
\end{itemize}

\section{Sensitivity to stemming}
We checked the sensitivity of our results to stemming by stripping plural forms from  our lower-cased words (the ones left after excluding function words and interjections), to then aggregate our counts by their stemmed counterpart, keep the top \SI{10000}{} forms and rerun our analysis. We reproduce Fig.~4 from the main text and show the outcome in Fig. S5. One can spot little differences in the extent of the cultural regions, but they are shown to be mainly the same. This shows that there is little sensitivity of our results to prior stemming of the corpus.

\clearpage

\begin{table}[hp!]
\centering
  \begin{tabular}{|| l | c | c | c ||} 
  \hline
  Year & Number of Tweets & Number of users & Number of tokens after filtering \\ [0.5ex] 
  \hline\hline
  2015 & $363 \times 10^6$ & $5.93 \times 10^6$ & $0.78 \times 10^9$ \\ 
  2016 & $805 \times 10^6$ & $6.80 \times 10^6$ & $1.78 \times 10^9$ \\
  2017 & $554 \times 10^6$ & $6.04 \times 10^6$ & $1.53 \times 10^9$ \\
  2018 & $524 \times 10^6$ & $5.49 \times 10^6$ & $1.71 \times 10^9$ \\
  2019 & $426 \times 10^6$ & $4.67 \times 10^6$ & $1.32 \times 10^9$ \\
  2020 & $446 \times 10^6$ & $3.58 \times 10^6$ & $1.29 \times 10^9$ \\
  2021 & $217 \times 10^6$ & $2.76 \times 10^6$ & $0.72 \times 10^9$ \\ [1ex] 
  \hline
  \end{tabular}
  \caption{Twitter dataset statistics. This table presents the number of Tweets and users in the US for every year between 2015 and 2021 included. The last column gives the total number of tokens that remain after we filter Tweets and users according to the Methods presented in the main text.}
\end{table}

\begin{figure}[hp]
\centering
  \includegraphics[width=0.8\textwidth]{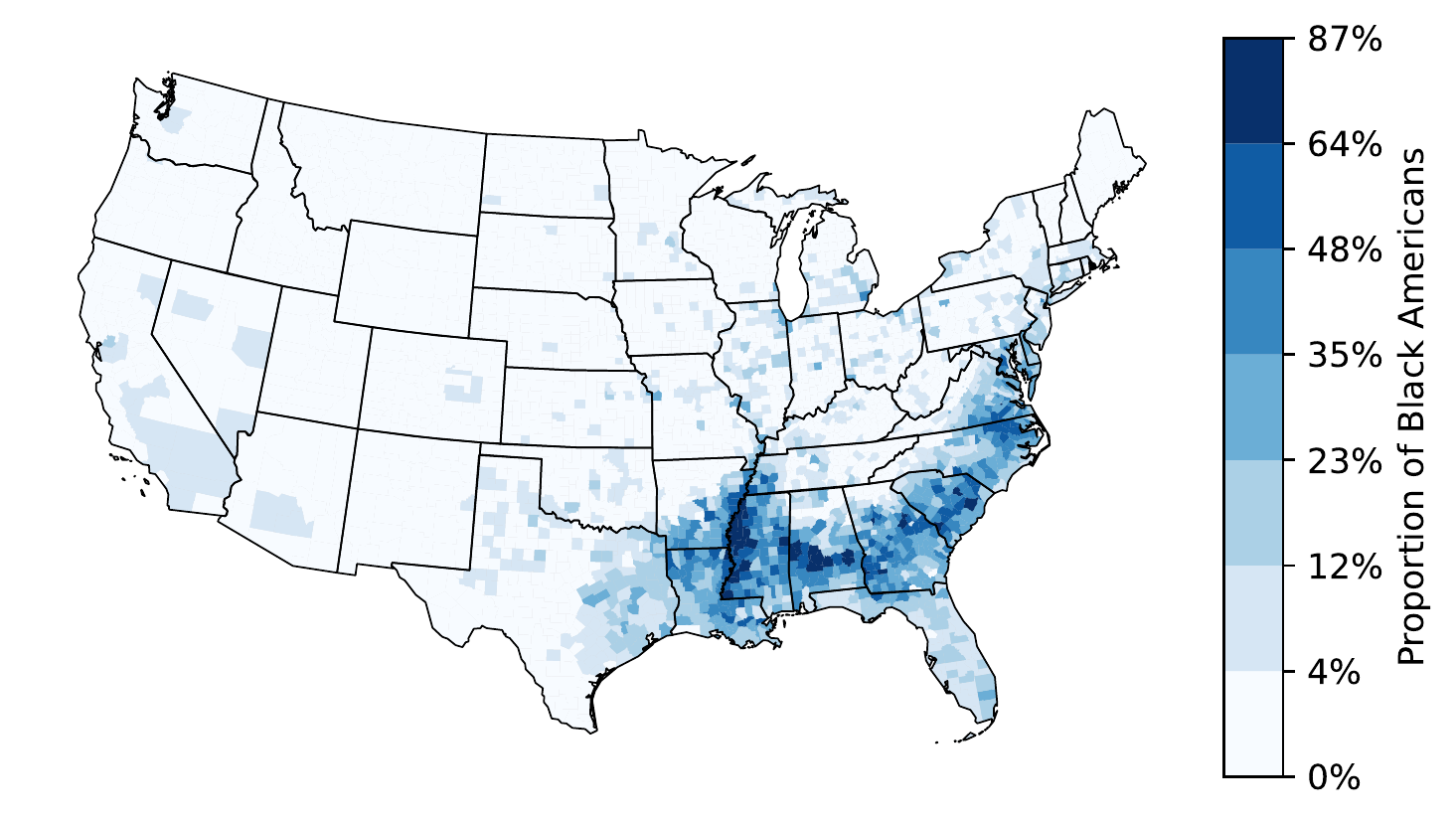}
  \caption{Proportion of Black Americans per county, as per the 2018 estimates of the US census.}
  \label{fig:census18_black_prop}
\end{figure}

\begin{figure}[hp]
\centering
  \includegraphics[width=\textwidth]{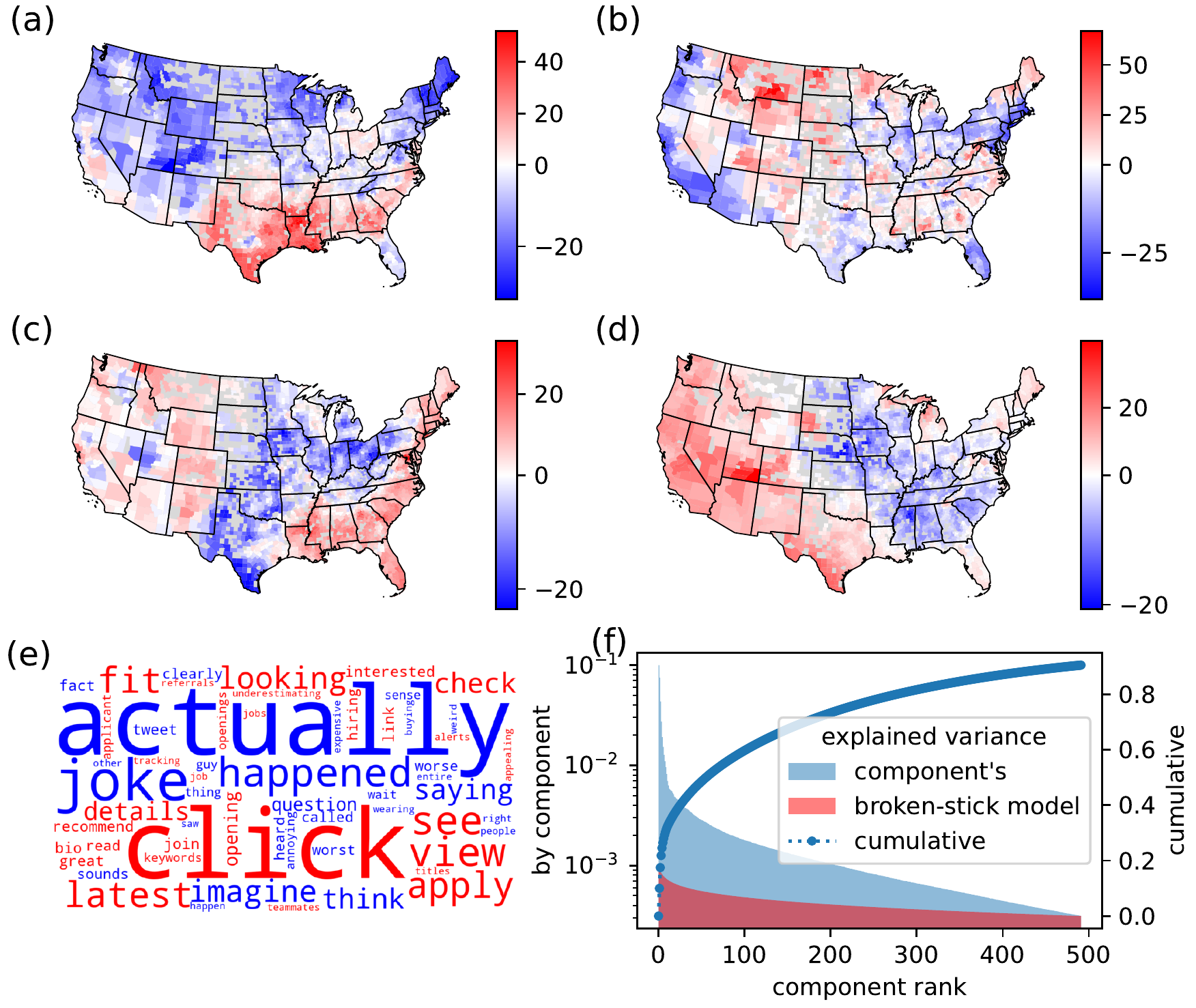}
  \caption{Result of the principal component analysis carried out on $G_i^*$ vectors calculated with a proximity matrix defined by considering 5 nearest neighbors. (a-d) Four maps show the projection of the data along the first four components, highlighting regional lexical variations. (e) Word cloud showing the words with strongest positive (red) and negative (blue) loadings for the second component, with each word's font size depending on its loading's absolute value. (f) Explained variance of the principal components compared to the broken-stick model on a logarithmic scale, which shows how the number of components to keep is selected at the first intersection of the two curves. We also plot the cumulative proportion of the variance explained by the components.}
  \label{fig:5knn}
\end{figure}

\begin{figure}[hp]
\centering
  \includegraphics[width=\textwidth]{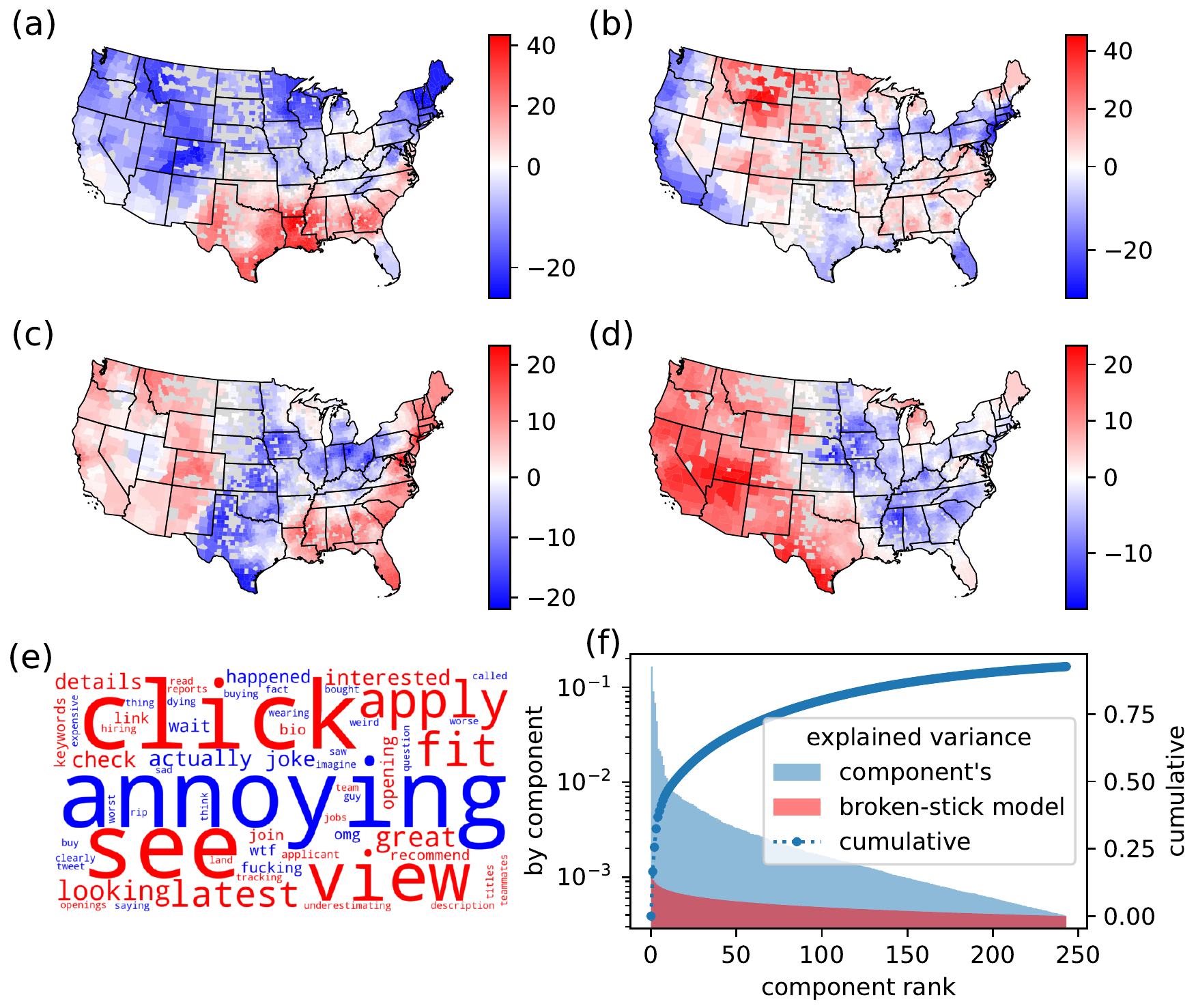}
  \caption{Result of the principal component analysis carried out on $G_i^*$ vectors calculated with a proximity matrix defined by considering 15 nearest neighbors. (a-d) Four maps show the projection of the data along the first four components, highlighting regional lexical variations. (e) Word cloud showing the words with strongest positive (red) and negative (blue) loadings for the second component, with each word's font size depending on its loading's absolute value. (f) Explained variance of the principal components compared to the broken-stick model on a logarithmic scale, which shows how the number of components to keep is selected at the first intersection of the two curves. We also plot the cumulative proportion of the variance explained by the components.}
  \label{fig:15knn}
\end{figure}
    
\begin{figure}[hp]
\centering
  \includegraphics[width=\textwidth]{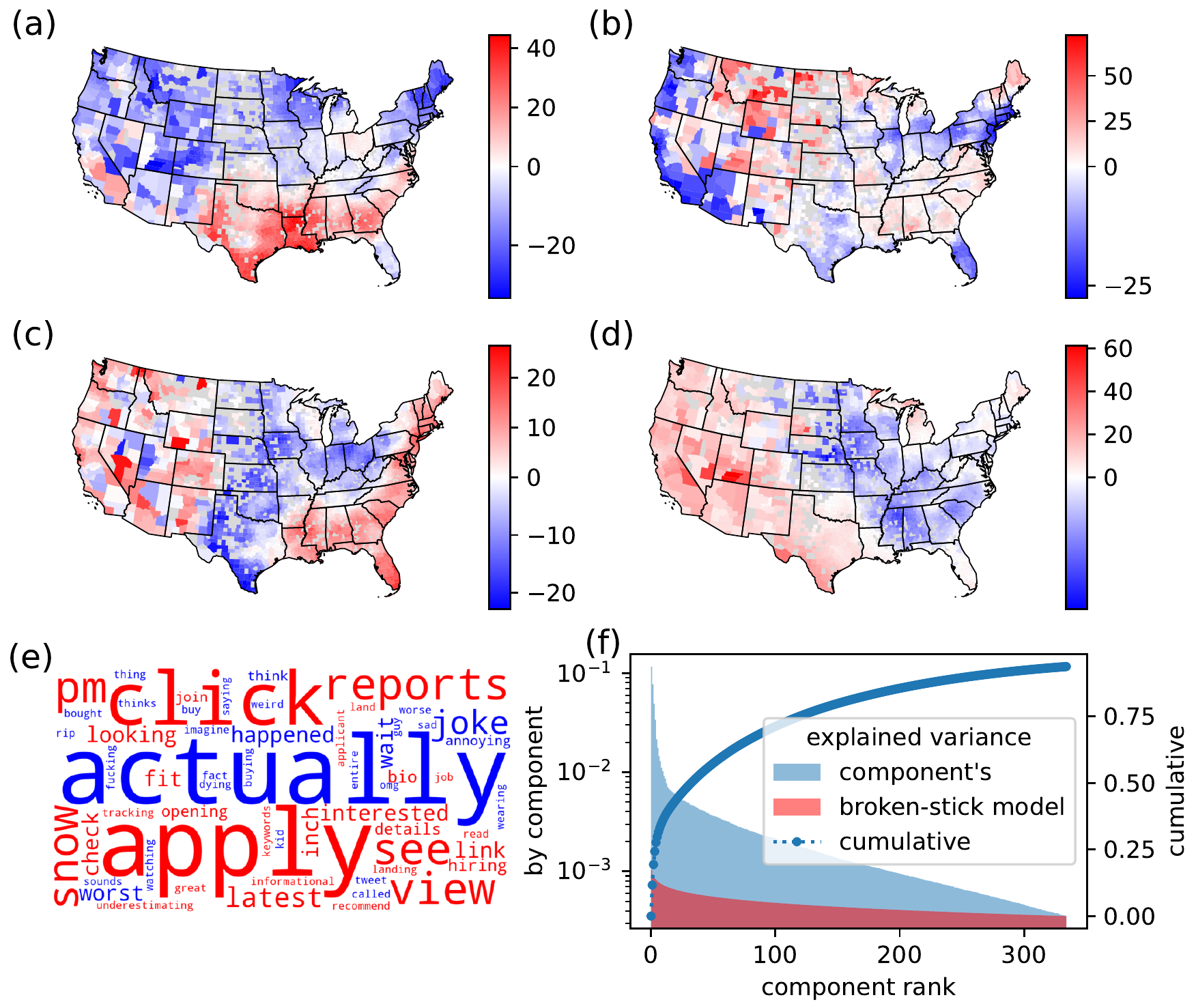}
  \caption{Result of the principal component analysis carried out on $G_i^*$ vectors calculated with a proximity matrix defined by a \SI{100}{\kilo \meter} distance band. (a-d) Four maps show the projection of the data along the first four components, highlighting regional lexical variations. (e) Word cloud showing the words with strongest positive (red) and negative (blue) loadings for the second component, with each word's font size depending on its loading's absolute value. (f) Explained variance of the principal components compared to the broken-stick model on a logarithmic scale, which shows how the number of components to keep is selected at the first intersection of the two curves. We also plot the cumulative proportion of the variance explained by the components.}
  \label{fig:DistanceBand}
\end{figure}
    
\begin{figure}[hp]
  \centering
    \includegraphics[width=0.9\textwidth]{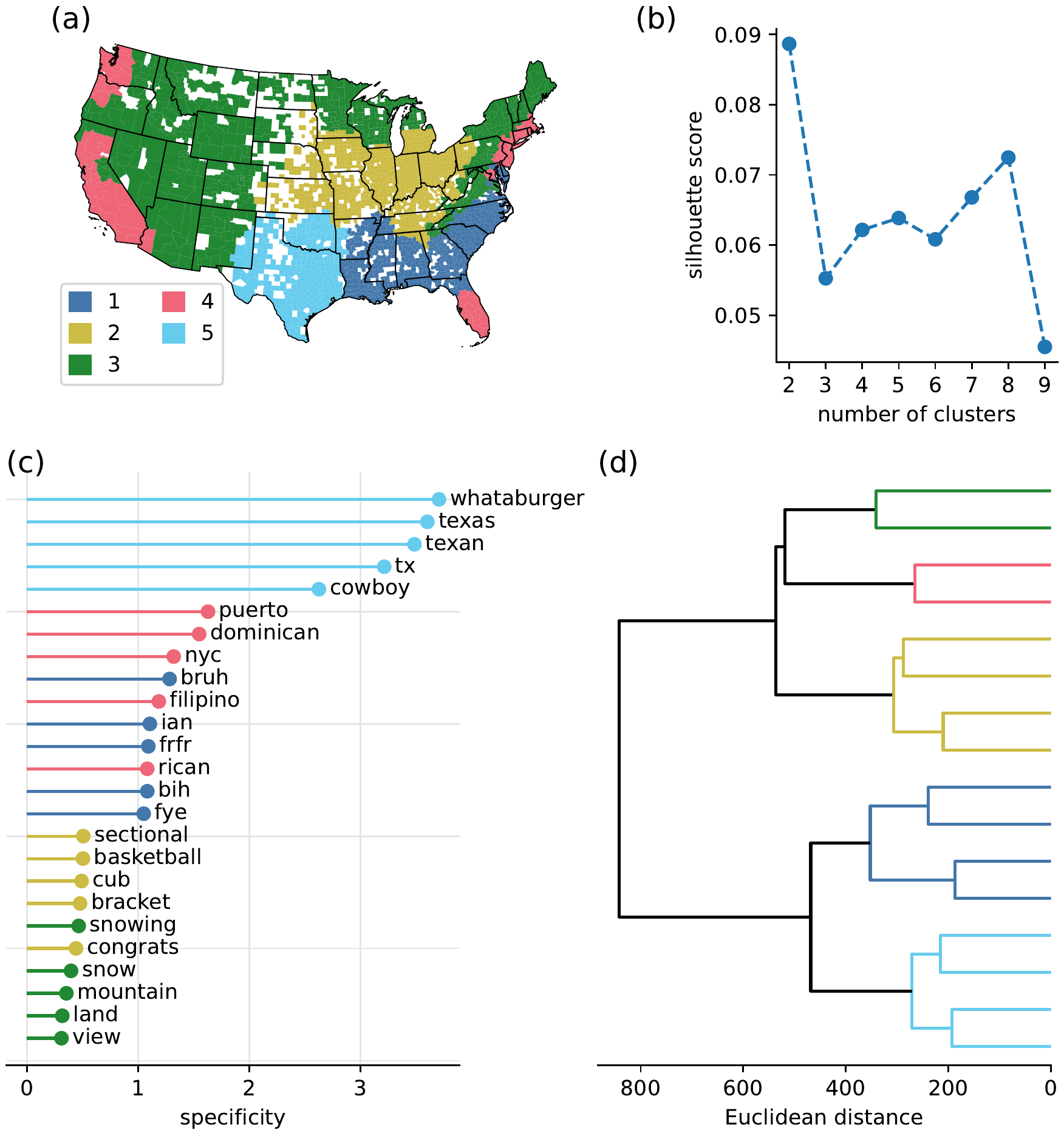}
    \caption{Cultural regions obtained when we keep the top \SI{10000}{} lemmas from our dataset. (a) Map of the five clusters obtained through hierarchical clustering, selected from a high value of (b) the mean Silhouette score. (c) Five most specific words for the five clusters shown in (a), along with their specificity values. (d) The dendrogram allows seeing which clusters are first joined if going to a higher level of clustering, and thus which ones are closer together. It clearly shows that the strongest division is the one between the North and the Southeast (excluding Florida) with further splittings as the cluster distance increases.}
    \label{fig:clust_with_stem}
\end{figure}

